\documentclass{egpubl}
\usepackage{sca2022}

 \SpecialIssuePaper   % uncomment for final version of Computer Graphics Forum, special issue
\usepackage[T1]{fontenc}
\usepackage{dfadobe}

\usepackage{bm}
\usepackage{amsfonts}
\usepackage{amssymb}
\usepackage{amsmath}
\usepackage{enumitem}
\biberVersion
\BibtexOrBiblatex
\usepackage[backend=bibtex,bibstyle=EG,citestyle=alphabetic,backref=true]{biblatex}
\electronicVersion
\PrintedOrElectronic
\ifpdf \usepackage[pdftex]{graphicx} \pdfcompresslevel=9
\else \usepackage[dvips]{graphicx} \fi

\usepackage{egweblnk}
\usepackage{color}
\definecolor{olive}{rgb}{0,0.4,0.2}
\definecolor{olivegreen}{rgb}{0.1,0.8,0.3}
\definecolor{mauve}{rgb}{0.48,0,0.72}

\newcommand{\na}[1]{{#1}\normalfont}

\def\etal{\textit{et~al.}}
\usepackage{accents}
\newcommand{\ubar}[1]{\underaccent{\bar}{#1}}

\title[Pose Representations for Deep Skeletal Animation]{Pose Representations for Deep Skeletal Animation}
\author[N. Andreou~\etal]
{\parbox{\textwidth}{\centering N. Andreou\thanks{nefeliandreou@outlook.com}$^{1,2}$\orcid{0000-0003-0341-6469}, A. Aristidou\thanks{a.aristidou@ieee.org}$^{1,2}$\orcid{0000-0001-7754-0791},  and Y. Chrysanthou$^{1,2}$\orcid{0000-0001-5136-8890}}
        \\
{\parbox{\textwidth}{\centering $^1$University of Cyprus, Nicosia, Cyprus,
         $^2$CYENS Centre of Excellence, Nicosia, Cyprus}}
}
\begin{document}

\teaser{
\centering
 \includegraphics[width=1\linewidth]{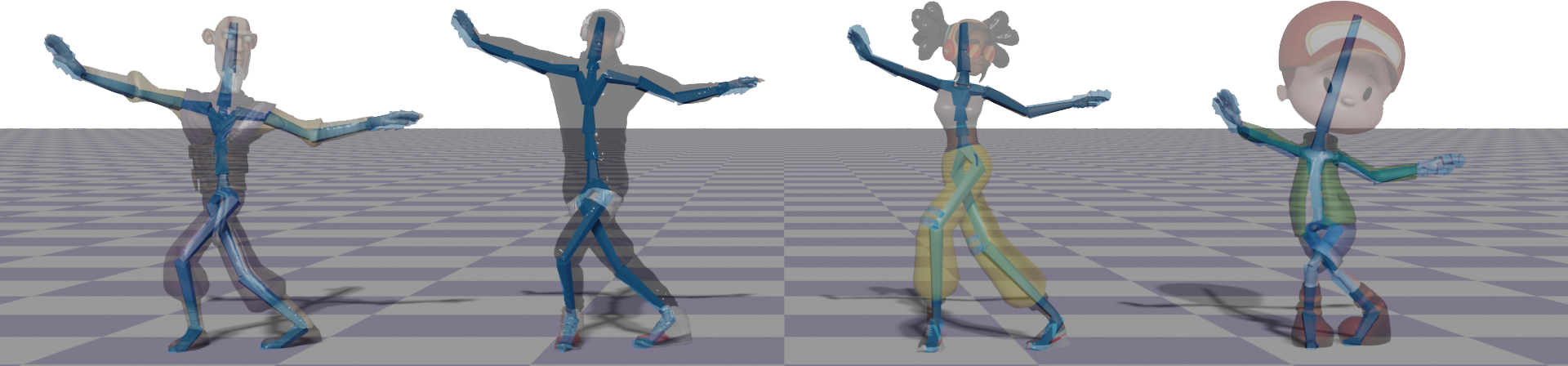}
 \caption{A fundamental component of motion modeling with deep learning is the pose parameterization. A suitable parameterization is one that holistically encodes the rotational and positional components. The dual quaternion formulation proposed in this work can encode these two components enabling a rich encoding that implicitly preserves the nuances and subtle variations in the motion of different characters.}
 \label{fig:teaser}
}
%  taking into consideration The motion varies according to the skeleton indicating that the skeletal hierarchy is a crucial factor in modeling motion. Most deep frameworks for synthesis retarget all motions to a universal skeleton for training. Our dual quaternion representation can encode skeletal information, thus facilitating learning and preserving the skeletal nuances in the motion of different characters.}

\maketitle
%-------------------------------------------------------------------------
\begin{abstract}
Data-driven skeletal animation relies on the existence of a suitable learning scheme, which can capture the rich context of motion.
% properly established model of motion, capable of describing its rich context.
However, commonly used motion representations often fail to accurately encode the full articulation of motion, or present artifacts. In this work, we address the fundamental problem of finding a robust pose representation for motion, suitable for deep skeletal animation, one that can better constrain poses and faithfully capture nuances correlated with skeletal characteristics. Our representation is based on dual quaternions, the mathematical abstractions with well-defined operations, which simultaneously encode rotational and positional orientation, enabling a rich
% hierarchy-aware
encoding, centered around the root. We demonstrate that our representation overcomes common motion artifacts, and assess its performance compared to other popular representations. We conduct an ablation study to evaluate the impact of various losses that can be incorporated during learning. Leveraging the fact that our representation implicitly encodes skeletal motion attributes, we train a network on a dataset comprising of skeletons with different proportions, without the need to retarget them first to a universal skeleton, which causes subtle motion elements to be missed. Qualitative results demonstrate the usefulness of the parameterization in skeleton-specific synthesis.
% We show that smooth and natural poses can be achieved, paving the way for fascinating applications.
%-------------------------------------------------------------------------
%  ACM CCS 1998
%  (see https://www.acm.org/publications/computing-classification-system/1998)
% \begin{classification} % according to https://www.acm.org/publications/computing-classification-system/1998
% \CCScat{Computer Graphics}{I.3.3}{Picture/Image Generation}{Line and curve generation}
% \end{classification}
%-------------------------------------------------------------------------

%\ccsdesc[500]{Computing methodologies~Motion capture}

\end{abstract}

% ------------------------------------------------------
% Keywords
% ------------------------------------------------------
%\keywords{Skeletal Hierarchy; Dual Quaternions; Neural Networks, Motion Modeling}

%% This command processes the author and affiliation and title information and builds the first part of the formatted document.
\maketitle

% ------------------------------------
% Main Document
% ------------------------------------

% \begin{figure}[h]
% \centering
%  \includegraphics[width=1\linewidth]{Figures/Teaser.png}
%  \caption{The motion varies according to the skeleton indicating that the skeletal hierarchy is a crucial factor in modeling motion. Most deep frameworks for synthesis retarget all motions to a universal skeleton for training. Our dual quaternion representation can encode skeletal information, thus facilitating learning and preserving the skeletal nuances in the motion of different characters.}
% \label{fig:teaser}
% \end{figure}

%\IEEEraisesectionheading{\section{Introduction}\label{sec:introduction}}
\section{Introduction}\label{sec:introduction}
Motion modeling is a fundamental task on the cornerstone of computer graphics and vision. A well established motion representation enables the synthesis of realistic human motion, satisfying constraints such as motion smoothness, continuity, and naturalism. Yet, computationally synthesizing human motion remains challenging, since motion is a highly stochastic process, governed by both intrinsic and extrinsic factors of which a harmonic balance should be achieved. % and the generated movements must be continuous, smooth, and expressive locally. 
Recent advances of machine learning and neural networks, along with the release of high-quality, large, diverse, and highly realistic motion capture databases, have shown promising results in reconstructing human motion from videos with high expressiveness, in synthesizing arbitrary movement with realism, and controlling articulated characters. % \yc{this [paragraph needs work. I would like to discuss it with you first.}]

The quality of the generated motion is highly influenced by two main factors: (a) the network architecture/design and learning scheme, and (b) the \na{input} pose representation. Over the last few years an enormous amount of research has been devoted to the architecture design and learning schema~\cite{Holden:2016,Fragkiadaki:2015,Barsoum:2017}.
% with previous works relying on convolutional~\cite{Holden:2016}, recurrent~\cite{Fragkiadaki:2015,Zhou:2018,Pavllo:2018,Wang:2019,Gopalakrishnan:2019}, phase-functioned~\cite{Holden:2017}, adversarial~\cite{Barsoum:2017}, fully-connected~\cite{Butepage:2017}, or Bayesian learning networks~\cite{Rui:2020}. 
However, less effort has been dedicated to the motion parameterization, even though it is of equal significance~\cite{Zhou:2019,Xiang:2020,Peretroukhin:2020}. Previous research has found that keeping the architecture fixed and experimenting with different representations affects the quality of the results~\cite{Pavllo:2018, Shi:2020}. In addition, it has been observed that common rotational representations still occasionally produce large errors when used for rotation regression tasks. This becomes more evident when the mapping between the network's embedding space and the original data is discontinuous, thus making it hard for the network to learn properly~\cite{Zhou:2019}. 
% We form our direction based on the fact that traditionally used representations such as Euler angles and quaternions are problematic because they are discontinuous~\cite{Zhou:2019,Xiang:2020}.
\na{As highlighted in a recent survey by Mourot~\etal~\cite{Mourot:2021}, a good pose representation for human motion modeling should accurately reflect the visual outcome while being suitable for optimisation (avoiding the error accumulation along the skeleton). It should capture both spatial and temporal correlations, leading to the extraction of informative patterns that serve a variety of tasks, such as rigging and skinning. The learnt patterns lay the grounds for inference, and determine the generalization ability of the model to different motions and skeletons. }

%Adequate pose parameterization is certainly a crucial component for motion generation with neural networks, as the networks learn the nature of motion and patterns in it from the given input representation.  %It has gained significant attention over the last few years, mainly because pose parameterization has been found to influence the quality of the produced motion e.g., rotational inconsistencies~\cite{Zhou:2019}. 
In early deep neural approaches, pose is represented using the Cartesian locations of joints, to ensure the continuity of the reconstructed motion~\cite{Holden:2016,Holden:2017,Cao:2018,Zhou:2018}. However, joint locations can only describe a small set of the full human motion articulation, resulting in ambiguities in motion reconstruction, related to the rotation on the roll axis. In addition, it is required to enforce priors (e.g., an Inverse Kinematics solver) about bone lengths to limit potential rig violations. In some later work, motion is represented as local joint rotations using 3D or 4D representations, such as Euler angles, axis-angle (exponential maps), or quaternions~\cite{Zhou:2019}. The space of the 3D/4D rotational representations, though, is not continuous causing singularities. Prediction error in crucial joints of the rig hierarchy is not properly encoded, resulting in error accentuation as we move to the end-effectors.
% yet positional losses are able to reflect this phenomenon, since an incorrect prediction at a crucial joint will be reflected on the error of all remaining joints of the hierarchical sequence until an end-effector is reached
To benefit from the trade-off between rotational and positional representations, recent works, such as~\cite{Pavllo:2018,Aberman:2020:Retargetting,Shi:2020}, model motion using joint local quaternions and employ a forward kinematics (FK) layer to recover or revise the corresponding joint positions. However, networks trained in such a way only rely on the positional information for supervision, hindering the rotational, which is a particular component in animation.
% \na{R4: clarify this sentence.}
% retrieving positions with FK does not enable the network to directly learn characteristics of rotation since these are hindered motion relating to the hierarchy of the skeleton for the examined characters.}

%  FK usually apply the loss only on the positional data hindering the rotational information which is important in animation
 
 % In a similar manner, a number of recent works decompose poses into the skeleton and the actual motion~\cite{Aberman:2020:Retargetting, Aberman:2019, Shi:2020}. The actual motion is inherently driven by the geometric properties of the skeleton, while components, adapt skeletal hierarchies during training, by adding a forward kinematics (FK) layer at the end to revise the skeleton. 

% \na{copied from Harvey:2020 A more informative way of representing the current offset to the target to would be to include positional-offsets in the representation. For this to be informative however, it would need to rely on character-local or global positions, which require FK. Although it is possible to perform FK inside the network at every step of generation, the backward pass during training becomes prohibitively slow justifying our use of root offset and rotational offsets only}

In addition, when training various models for motion related tasks, previous works suggest to retarget all motions in the database into a universal skeleton configuration, in order to allow for generalizable character and setup agnostic experiments. This is because databases consist of motion captured using actors, where each performer's body structure varies in terms of bone lengths and proportions, shaping their motion traits. However, such a requirement limits the ability of networks to support expressive motion attributes and nuances that were originally performed by actors with different skeletal structure, and were lost in the retargeting process. 

%We form our methodology based on the fact that skeletal information assists learning. you need to be more specific. Why and when Experiments performed by ~\citet{Pavllo:2018} indicate that input representation as well as the positional loss function employed during training are particularly impactful \na{to capture large errors in crucial joints of the rig hierarchy.} \Andreas{again, what loss and why}. 

In this paper, we tackle the fundamental problem of representing poses, using
%propose
a hierarchy-aware representation that uses dual quaternions as the mathematical framework.
%for motion parameterization. %\yc{I am a bit confused here. Is it hierarchy-aware because it uses positional and rotational info, or is it because of the way the distances are computed - all from the hip?}
The main premise of our method is that dual quaternions allow us to define poses in a root-centered manner, implicitly encoding skeletal information that can be leveraged by the network. Dual quaternions provide a unified, elegant, and compact representation that incorporates rotational and translational information in the form of orthogonal quaternions~\cite{Kenwright:2012}. They carefully encode the aforementioned elements of human motion in one component, and parse it as an input to the network, so that the model benefits from both parameters, % We form dual quaternions so as they sequentially encode the transformations of joints along the hierarchy \yc{this sentence is not clear to me}, allowing us to directly extract positional information, and employ a positional loss, %The proposed motion representation is capable of encoding additional knowledge for the hierarchical structure compared to traditional techniques,
thus leading to more constrained learning. In this way \na{we can directly extract the positional information from the representation avoiding the need for external kinematic operations such as FK. Then we can use this information as a loss, similar to recent works e.g.,~\cite{Shi:2020,Pavllo:2018,Harvey:2020}}.
Our main contribution is the exploitation of a hierarchy-aware pose motion representation based on the properties of dual quaternions, as well as the assessment of currently used inputs for deep skeletal animation. The proposed representation %Such a representation 
allows to infer both rotational and positional information directly, while it encodes the correlations between joints and limbs along the structure of the rig. We integrate several constraints, in the form of training losses, to better penalize the errors in crucial joints of the hierarchy; the losses operate directly on the representation, eliminating the need for external kinematic calculations during training. To the best of our knowledge, the dual quaternion properties have not been explored before in the context of human motion modeling with deep neural frameworks. In particular, we hypothesize that learning from such a representation may enable the network to better exploit skeletal-nuances of motion, enabling the generation of stable motion on with less training. This representation can be used for a variety of tasks, such as motion prediction or synthesis, motion retargeting, motion reconstruction, and it is not tuned towards specific motion patterns. We demonstrate the effectiveness and practicality of the proposed representation using two well-known network architectures which focus on the fundamental application of motion synthesis: the auto-conditioned Recurrent Neural Network (acRNN) of Zhou~\etal~\cite{Zhou:2018}, and the QuaterNet developed by Pavllo~\etal~\cite{Pavllo:2018}. In particular, we perform both short-term prediction and long-term synthesis, we examine the relevance of the proposed losses for each architecture through an ablation study, and show that depending on the task, each loss serves a purpose. Finally, we show that the proposed parameterization can be used for skeleton-specific synthesis, where the characters during training can have varying proportions (see Figure~\ref{fig:teaser}).

\section{Related Work}
\label{section: Related Work}
%Machine learning models are important in modern graphics since computers can now learn from data. The game has change due to the recent development in computational power, and the large data availability that is publicly accessible for training these models. In that manner, 
In recent years, various deep learning models have been developed to accomplish complex tasks in character animation, including motion reconstruction~\cite{Cao:2018,Shi:2020}, action recognition~\cite{Du:2015}, motion synthesis~\cite{Holden:2016,Holden:2017,Zhou:2018}, prediction~\cite{Fragkiadaki:2015,Wang:2018}, style transfer~\cite{Aberman:2020,Aristidou:2020:MIG,Smith:2019}, motion retargeting~\cite{Aberman:2020:Retargetting,Du:2019} etc. 
%As those models are iteratively exposed to new data, they are capable to independently adapt; they learn from previous computations to produce reliable, repeatable decisions and results. 
Over the last few years, a number of different architectures and learning schemes have been introduced to model human articulation, including convolutional~\cite{Holden:2016,Butepage:2017}, recurrent~\cite{Fragkiadaki:2015,Zhou:2018,Martinez:2017,Jain:2016,Wang:2021:STRNN}, phase-functioned~\cite{Holden:2017, Starke:2020b}, or adversarial~\cite{Barsoum:2017,Wang:2018,Wang:2019} networks. Furthermore, generative models based on normalizing flows~\cite{Henter:2020} and VAEs~\cite{Ling:2020} are becoming popular since they can generate diverse motions, and have proven to be effective for a variety of tasks such as control or planning. However, despite the progress made on the design of advanced networks to model the high frequencies of motion, the proposed architectures still depend on simplified pose representations (e.g., joint locations or local rotations) which cannot encode the full articulation of human motion. Even with abundant weights and extensive training, they still occasionally produce big errors. In previous work, the efficiency of deep learning approaches has been found to be highly dependent upon the quality of the dataset and the representation used~\cite{Xiang:2020,Zhou:2019,Ma:2019}, due to the fact that the network learns motion patterns from the data. A rich encoding of motion, based on a concrete mathematical model, encourages the network to properly disentangle complex characteristics of motion which are present in the data. % such frameworks automatically identify the features of a given data. 

\subsection{Motion parameterization}

Human motion is often modelled as a sequence of skeletal states expressed in terms of the 3D orientation of the bones (angular representations) or the 3D coordinates of joints (positional representations). Each parameterization has its own benefits and limitations, which we briefly outline below. As pointed out in a recent survey, the parameterization of input and output guide the network to retain specific features~\cite{Mourot:2021}. We argue that a representation that unifies positional and rotational information can achieve the maximum potential in character animation.

\textit{Positional Pose Representations:} In early machine learning approaches, large amounts of work have been devoted to the development of deep learning networks that use 3D joint positions \cite{Holden:2016, Fragkiadaki:2015,Zhou:2018}, since minimizing the 3D position errors ensures prediction of correct joint locations, and maintains that mispredictions on crucial joints are taken into consideration. Positional data, however, comes at several costs. First, for a pair of fixed consecutive 3D positions there exist multiple limb rotations, which can be recovered using Inverse Kinematics (IK). This leads to ambiguities on the rotation of each limb on its roll axis. In addition, such representation lacks the benefit of the parameterized skeleton. If the generated motion is to be applied on a different skeleton, post-processing is required to secure that the skeleton constraints are satisfied, i.e., bone constraints or motions within the articulation range. 
% Furthermore, there is information loss during the conversion of 3D positional data to the corresponding 3D motion representation, caused by the lack of information on the longitudinal rotation component. 
Consequently, positional data fails to describe the full range of human motion articulation, thus does not suffice to uniquely recover human characters. Considering the aforementioned reasons, positional representations are less \na{suitable for graphics and animation applications}. 
% popular in the animation community yet remain common in the computer vision community.

%J On the other hand, 3D positional per joint information can be used as input. Each parameterization has its own benefits and limitations, which we briefly outline below. We argue that we can encode both representations in a unified way to achieve the maximum potential.

\textit{Angular Pose Representations:} Rotational representations belong to the rotation group $SO(3)$, and can be expressed using various parameterizations e.g. Euler angles, axis-angle, quaternions, to mention a few. The main benefit of rotational representations over positional is that they allow for a parameterized skeleton~\cite{Aberman:2020:Retargetting,Shi:2020}, thus avoid prediction errors related to bone stretching or motion outside the articulation range. % The limitations vary according to the exact representation used to model joint rotations, as well as the loss accompanying the network. 
Euler angles are the most intuitive representation, and have been widely explored in previous research~\cite{Aristidou:2018,Zhou:2019}. They can be used to describe the orientation of a rigid body as successive rotations relative to a fixed coordinate system $x,y,z$. %Euler angles are one of the most convenient representations as they can be extracted directly from motion files such as BVH files. 
However, Euler angles come with many limitations: they are discontinuous and non-unique in the sense that for a particular orientation, $\theta$ and $\theta + 2k\pi$ for $k \in \mathbb{Z}$ represents the exact same rotation, causing learning problems. In addition, they are prone to the Gimbal lock effect. To mitigate the issues caused by Euler angles, exponential maps (axis-angle) representations have been adopted~\cite{Martinez:2017,Wang:2018,Henter:2020}. Axis-angle representations have a major drawback, that is they cannot express composition of rotations~\cite{Grassia:1998}, thus cannot be used for hierarchical modeling. Another way of overcoming the discontinuity issue of Euler angles is to represent each angle $\theta$ using a 2D vector, $[\text{cos} \theta,\text{ sin} \theta]$, or equivalently a unit complex number $a + bi$. To perform smooth regression the constraint $a^2 + b^2 = 1$ should be maintained. This approach, doubles the number of parameters that need to be predicted, and does not benefit from incorporating positional information.

On the other hand, \na{many works use unit quaternions to encode joint rotations when training deep networks in character animation, demonstrating satisfactory results in several tasks, such as motion reconstruction from video~\cite{Shi:2020}, style transfer~\cite{Dong:2017}, motion retargeting~\cite{Villegas:2018}, motion synthesis and prediction~\cite{Pavllo:2018,Aristidou:2022}.}
% been established as the most common encoding of joint rotations for training deep networks in character animation, demonstrating satisfactory results in several tasks, such as motion reconstruction from video~\cite{Shi:2020}, style transfer~\cite{Dong:2017}, motion retargeting~\cite{Villegas:2018}, motion synthesis and prediction~\cite{Pavllo:2018,Aristidou:2022}.
However, despite the growing popularity of unit quaternions, Zhou~\etal~\cite{Zhou:2019} and Xiang~\etal~\cite{Xiang:2020} have recently demonstrated experimentally that the quaternion's space is still not continuous. More specifically, they showed that motion continuity cannot be achieved for a space with less than four dimensions ($\mathbb{R}^4$), and proved that % Experimenting with a multi-layer perceptron for a sanity test and an Inverse Kinematics task, ~\citet{Zhou:2019} showed that 
neural networks can learn better from continuous representations. Zhou~\etal~\cite{Zhou:2019} pointed out that a rotation representation can be made continuous using the identity mapping, which would result in $n \times n$ sized matrices ($n$ is the dimension of rotations), which may not only be excessive, but also still require orthogonalization in mapping from the representation to the original space. Thus, they proposed to perform an orthogonalization in the representation itself, resulting in a 6 dimensional parameterization (ortho6D). This representation has been widely adopted in recent works~\cite{Zhang:2018,Ling:2020,Petrovich:2021}. However, it does not encode positional information, while a hierarchical modeling with FK would require conversion to transformation matrices, in order to perform the sequential encoding of orientations along the rig. %Even though this representation is more robust than Euler angles, it has been found to lead to self-rotations as well as minor global shaking when employed in the MotioNet ~\cite{Shi:2020}.

\begin{figure}
    \centering
    \includegraphics[scale=0.4]{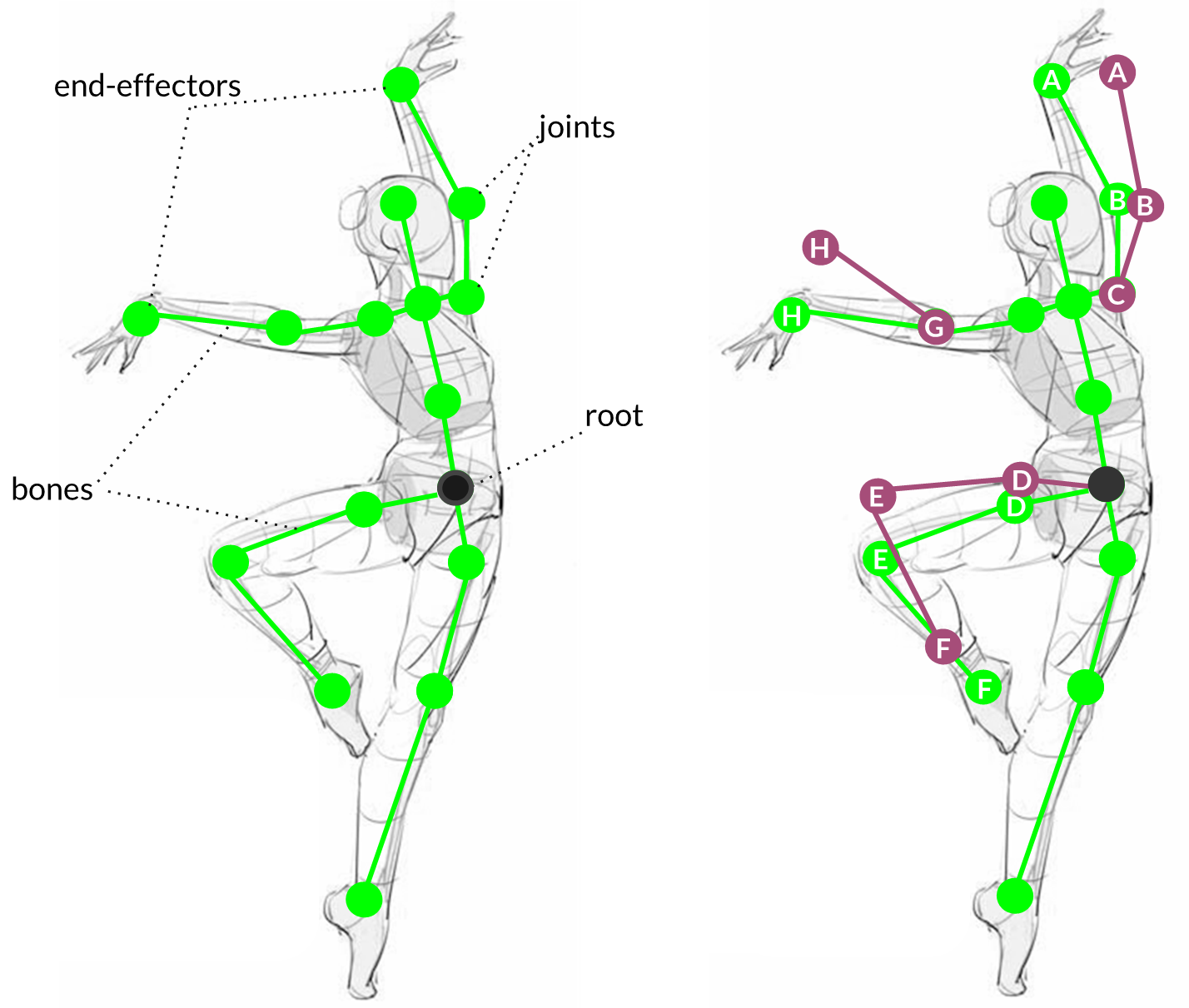}
    \caption{Error accumulation along kinematic chain. We can see that a small local rotation error in either joint C or D drastically affects the position of joints A,B,E,F. Errors on joint G are less crucial as it is not at the base of the hierarchical chain and its transformation only affects joint H. Taking a weighted average of the local joint rotation errors does not reflect that a wrong prediction in joint D has a larger impact on the resulting motion, than joint G.}
    \label{fig:error_accum}
\end{figure}

 In recent works, networks which rely on rotational parameterizations often integrate an additional FK layer, and are paired with a positional loss to further constrain motion~\cite{Pavllo:2018, Aberman:2020}. It has been observed that this is a necessary step for designing accurate learning schemes since small prediction errors in certain joint's rotations fail to be properly encoded in losses which average local rotation errors, yet drastically impact the positional error. In practice, the prediction error is accumulated along the hierarchy, as illustrated in Figure~\ref{fig:error_accum}. 
%
% \subsubsection{Euler Angles}

\textit{Hybrid Representations:}
Several works proposed the use of hybrid representations in order to get the best of both worlds. It has been demonstrated that \na{additional information in the parameterization, such as velocities, }benefits the learning process. Some hybrid representations include joint orientations with joint positions~\cite{Lee:2018}, positions and velocities~\cite{Holden:2017,Zhang:2018,Starke:2020b,Starke:2019}, and joint positions with linear and angular velocities~\cite{Holden:2020,Li:2020}. In contrast, our parameterization is equally expressive and combines the benefits of positional and angular information in a unified entity, but does not include redundant information.

\subsection{Dual Quaternions in Character Animation}
Dual quaternions are powerful algebraic constructs which have been successfully used in a number of domains. Their power lies in the fact that they can efficiently represent rotations and translations in a unified way, while their well-defined algebraic operations such as multiplication, make them appealing for hierarchical modeling. They have been used extensively in computer vision and robotics~\cite{Goddard:1998,Daniilidis:1999,Torsello:2011}, and more recently have been adopted in character animation.
%~\cite{Goddard:1998} use dual quaternions to represent 3D transformation and perform pose and motion estimation. \citet{Daniilidis:1999} use dual quaternions to facilitate hand-eye calibration by simultaneously solving for the rotation and translation components. \cite{Torsello:2011} developed a motion registration algorithm which represents motions using dual quaternions.
For instance, Kavan~\etal~\cite{Ladislav:2008} apply dual quaternions for skeletal skinning and achieve faster execution times than more traditional skinning methods. Kenwright~\cite{Kenwright:2012} demonstrates the superiority of dual quaternions over traditionally used transformation matrices, while Vemulapalli~\etal~\cite{Vemulapalli:2016} examine various skeletal representations, for the purpose of human action recognition. Comparisons between representations that use joint rotations and joint translations separately, indicate that the newly proposed family of skeletal representations achieve higher performance for the task of action recognition on a wide range of datasets. It is worth noting that the work of Kenwright~\cite{Kenwright:2012} demonstrates that the well formulated algebraic operations of dual quaternion make them ideal candidates to represent rigid transformations of hierarchical chains, with both performance gains compared to rotation matrices and computational benefits when calculating angular and linear differences.

\na{In this paper, dual quaternions are used for the first time to parameterize skeletal motion in a deep learning framework.
% with neural networks to accomplish skeletal animation tasks such as motion prediction. 
%A recent work exploits the use of dual quaternions for rigid body movement prediction using neural networks~\cite{Schwung:2021}. 
In contrast to previous works that have designed a neural network architecture where each node corresponds to a dual quaternion~\cite{Schwung:2021},  %, who design a neural network architecture where each node corresponds to a dual quaternion rather than a scalar value, 
we use dual quaternions to model complex movement of the skeleton which has multiple connected joints. To handle the skeletal structure and introduce an inductive bias in the learning process, we express the representation of each joint in current coordinates. Thus, in our work, dual quaternions encode the hierarchical nature of a rig, enabling the integration of losses that take into consideration the rotation and displacement components. This formulation assists the learning, leading to more realistic synthesis.}

% \na{focus more on the learning}
%for faster convergence. (i more accurate/natural modelling?)

\section{Mathematical Representation}
\label{section: Mathematical Representation}

%In this work, we utilize dual quaternions. 
A dual quaternion $\mathbf{\ubar{q}}$ can be represented as an ensemble of ordinary quaternions $\mathbf{q}_r$ and $\mathbf{q}_d$, in the form $\mathbf{q}_r +  \mathbf{q}_d\varepsilon $, where  $\varepsilon$ is the dual unit, satisfying the relation $\varepsilon^2= 0$. The first quaternion describes the rotation. The second quaternion, $\mathbf{q}_d$ encodes translational information which we will define in Section \ref{sec:Dual Quaternion Motion Representation}. Each quaternion is characterized by 4 DOF making the dual quaternion an 8D representation, which can be considered as an 8-tuple of real values~\cite{Kavan:2005}. A unit dual quaternion can be interpreted as a manifold in the 8-dimensional Euclidean space~\cite{Ladislav:2008}.

Dual quaternions allow for convenient mappings from and to other representations which are currently used in the literature, allowing for effortless integration into current architectures. To reduce ambiguity we establish the following notation which will be used throughout this paper:
\begin{table}[h]
\centering
\begin{tabular}{cc}
$\mathbf{q}$ quaternion & $\mathbf{\ubar{q}}$ dual quaternion    \\
$\mathbf{\hat q}$ unit quaternion & $\mathbf{\ubar{\hat q}}$ unit dual quaternion\\
$\mathbf{q^*}$ quaternion conjugate & $\mathbf{\ubar{q}^*}$ dual quaternion conjugate  \\
\end{tabular}
\end{table}

\subsection{Basic Algebraic Properties}
\label{subsection: Basic Algebraic Properties}
Dual quaternions have well-defined algebraic operations which we list below. To put in context, in the following sections we elaborate on the construction of dual quaternions from other representations, that is, dual quaternion to and from rotation and 3D position. \\[0.1cm]
\textbf{Multiplication:} Suppose that $\mathbf{\ubar{q}}_1, \mathbf{\ubar{q}}_2$ are dual quaternions of the form $\mathbf{\ubar{q}}_1 = \mathbf{q}_{r1} + \mathbf{q}_{d1} \varepsilon$, $\mathbf{\ubar{q}_2} = \mathbf{q}_{r2} + \mathbf{q}_{d2} \varepsilon$ then the multiplication operation is defined as:
\begin{equation}
\mathbf{\ubar{q}}_1\mathbf{\ubar{q}}_2 = \mathbf{q}_{r1}\mathbf{q}_{r2} + (\mathbf{q}_{r1}\mathbf{q}_{d2} + \mathbf{q}_{d1}\mathbf{q}_{r2})\varepsilon.
\end{equation}
\textbf{Conjugate:} The conjugate of $\mathbf{\ubar{q}}$, i.e. $\mathbf{\ubar{q}^*} = \mathbf{q}^*_{r} + \mathbf{q}^*_{d}\varepsilon$ \\[0.1cm]
\textbf{Magnitude:} The magnitude of a dual quaternion is given by: 
\begin{equation}
||\mathbf{\ubar{q}}|| = \sqrt{\mathbf{\ubar{q}}\mathbf{\ubar{q}^*}} = ||\mathbf{q_r} || + \varepsilon \frac{<\mathbf{q}_r,\mathbf{q}_d>}{ ||\mathbf{q}_r ||}.   
\label{eq:norm}
\end{equation} 
Here $<a,b>$ denotes the dot product between the real valued vectors $a$ and $b$. \\[0.1cm]
\textbf{Unitary condition:} A unit dual quaternion should satisfy the following two conditions:
\begin{enumerate}
    \item the real part $\mathbf{q}_r$ must be a unit quaternion, i.e. $||\mathbf{q}_r ||=1$
    \item the real part must be orthogonal to the dual part, i.e. 
    \begin{equation}
       \mathbf{q}^*_r\mathbf{q}_d + \mathbf{q}^*_d\mathbf{q}_r =0
   \label{eq:unit}
   \end{equation} 
\end{enumerate} 
For long multiplication chains, it might be necessary to re-normalize a dual quaternion to mend drift. To make a dual quaternion unit, we can divide it by its magnitude. \\[0.1cm]
% From Eq.~\ref{eq:norm}, provided that the real and dual quaternions are orthogonal to each-other, normalization breaks down to normalizing the real part, and dividing the dual part by the magnitude of the real. \\[0.1cm]
%
\textbf{Inverse:} The inverse of a unit dual quaternion is equal to its conjugate, i.e. $\mathbf{\ubar{\hat{q}}^{-1}} = \mathbf{\ubar{\hat{q}}^*}$.

Dual quaternions can represent transformations. A pure rotation can be represented by a unit dual quaternion with zero dual part, namely $\ubar{\mathbf{q}} = (w_r + x_r\mathbf{i} + y_r \mathbf{j} + z_r\mathbf{k}) + \varepsilon(0 +0\mathbf{i} +0\mathbf{j} + 0\mathbf{k} )$, whereas a pure displacement by $x,y,z$ units on each of the unit axes respectively can be represented by a dual quaternion with the identity quaternion as real part, i.e. $\ubar{\mathbf{q}} = (1 + 0\mathbf{i} + 0 \mathbf{j} + 0\mathbf{k}) + \varepsilon(0 +\frac{x}{2}\mathbf{i} +\frac{y}{2}\mathbf{j} + \frac{z}{2}\mathbf{k})$. The operation $\ubar{\mathbf{\hat{q}}}\mathbf{p}\ubar{\mathbf{\hat{q}^*}}$ can be used to transform a point $\mathbf{p}$ inserted in a quaternion.

\subsection{Dual Quaternion from 3D rotation and 3D position}
\label{subsec:fromquatpos}

The rotation quaternion can be defined as the combination of Euler rotations along each of the $x,y,z$ axes. The unit quaternion that corresponds to a rotation of angle $\theta$ around axis $\mathbf{n}$ is given by:
\begin{equation}
    \mathbf{\hat{q}}_r = (\text{cos} \frac{\theta}{2}, \mathbf{n} \text{ sin}\frac{\theta}{2})
\end{equation}
Therefore, a rotation of $\alpha$ radians along the $x$-axis corresponds to the quaternion $\mathbf{\hat{q}_{rx}} =\text{cos}\frac{\alpha}{2} + \text{sin}\frac{\alpha}{2} \mathbf{i} + 0\mathbf{j} + 0\mathbf{k}$. Similarly for rotation $\beta$ along the $y$-axis and $\gamma$ along the $z$-axis the corresponding quaternions are $\mathbf{\hat{q}_{ry}} =\text{cos}\frac{\alpha}{2} + 0 \mathbf{i} + \text{sin}\frac{\beta}{2}\mathbf{j} + 0\mathbf{k}$ and $\mathbf{\hat{q}_{rz}} =\text{cos}\frac{\alpha}{2} + 0 \mathbf{i} + 0\mathbf{j} + \text{sin}\frac{\gamma}{2}\mathbf{k}$. Then, we multiply the quaternions based on the order of rotation. For example, if the order of rotation is $zyx$, the quaternion $\mathbf{\hat{q}_r}$ describing the resulting orientation can be obtained by: 
\begin{equation}
    \mathbf{\hat{q}}_r = \mathbf{\hat{q}}_{rz}\mathbf{\hat{q}}_{ry}\mathbf{\hat{q}}_{rx}.
    %= 
    %(\text{cos}\frac{\alpha}{2} \text{cos}\frac{\beta}{2} \text{cos}\frac{\gamma}{2} +  \text{sin}\frac{\alpha}{2}\text{sin}\frac{\beta}{2}\text{sin}\frac{\gamma}{2}) + \\ (\text{sin}\frac{\alpha}{2}\text{cos}\frac{\beta}{2}\text{cos}\frac{\gamma}{2} - \text{cos}\frac{\alpha}{2}\text{sin}\frac{\beta}{2}\text{sin}\frac{\gamma}{2})\mathbf{i} + ()
\end{equation} 
The dual part, $\mathbf{q}_d$ can be constructed as:
\begin{equation}
    \mathbf{q}_d = \na{\frac{1}{2}\mathbf{q}_t\mathbf{\hat{q}}_r}
    \label{eq:displacement}
\end{equation}
where $\mathbf{q}_t = 0 + x\mathbf{i} + y\mathbf{j} + z\mathbf{k}$, where $x,y,z$ denotes 3D displacement in Cartesian coordinates.

% \subsection{Dual Quaternion from homogeneous matrix}
% Similarly to what has been described before, we can construct a dual quaternion from a homogeneous matrix. Using the routine for the conversion of transformation matrices to quaternions we can construct $\mathbf{q}_r$. \na{shall I explain how we get to rotation matrices from quaternions}
% To construct the quaternion $\mathbf{q_d}$, we utilise the final column of the homogeneous matrix which provides the position vector. We can utilise the given $x,y,z$ values to construct a quaternion of the form $\mathbf{q}_t = 0 + x\mathbf{i} + y\mathbf{j} + z\mathbf{k}$ and use it to construct $\mathbf{q}_d = \frac{1}{2}\mathbf{q}_t.\mathbf{q}_r$ where $.$ denotes quaternion multiplication as defined in \cite{Kenwright:2012}.

% For a transformation $\mathbf{\hat{q}_r}$, the equivalent transformation matrix $R$ is 
% \begin{equation}
% R = \begin{bmatrix}
% 1-2(y_r^2+z_r^2) & 2(x_ry_r-z_rw_r)& 2(x_rz_r+y_rw_r) \\
% 2(x_ry_r+z_rw_r) & 1-2(x_r^2 + z_r^2) & 2(y_rz_r-x_rw_r) \\
% 2(x_rz_r - y_rw_r) & 2(y_rz_r+x_rw_r) & 1-2(x_r^2+y_r^2)
% \end{bmatrix}
% \end{equation}
% Details relating to the recovery of the rotational information are omitted in this paper since they are straightforward considering that one can use the rotational quaternion $\mathbf{q}_r$ and apply the usual quaternion transformations. \Andreas{NO,we have to give all the details. Papers should be self-consistent.}

\subsection{Recovering the 3D rotation and translation in Cartesian coordinates}
Apart from constructing the dual quaternions, we make sure that we can recover the rotational component in Euler angles, which we then use to construct a BVH file. Given a unit dual quaternion $\mathbf{\ubar{\hat{q}}}$, we can recover the rotation from the rotational part. For example, assuming that the order of rotation is $zyx$ we can obtain angles $\alpha,\beta$ and $\gamma$, which correspond to rotations along the $x,y,z$ axes by:
\begin{equation}
\begin{split}
\alpha & =\text{arctan}\bigg(\frac{2 (w_rx_r + y_rz_r)}{1 - 2(x_r^2 + y_r^2)}\bigg) \\
 \beta & = \text{arcsin}\big({2(w_ry_r - z_rx_r)}\big)\\
 \gamma & = \text{arctan}\bigg(\frac{2(w_rz_r+x_ry_r)}{1-2(y_r^2+z_r^2)}\bigg)
\end{split}
\label{eq:rot}
\end{equation}
The translation component in Cartesian coordinates can be obtained using the displacement quaternion,
\begin{equation}2 \mathbf{q}_d\mathbf{\hat{q}}_r^*,
\label{eq:translation}
\end{equation} where $\mathbf{\hat{q}}_r^*$ denotes the unit quaternion conjugate of the rotational quaternion. The Cartesian coordinates $x,y,z$ are the coefficients of the unit vectors $\mathbf{i},\mathbf{j},\mathbf{k}$.

For more details on quaternions and dual quaternions the reader is referred to the works of Dam~\etal~\cite{Dam:1998} or Kenwright~\cite{Kenwright:2012}.

\section{Motion Representation}
\label{section: Motion Representation}

In skeletal animation, the rig is represented as a hierarchical graph consisting of interconnected bones (edges), usually referred to as joints and end-effectors. Joints are connected with a parent/child relation. Each joint can be subject to an affine transformation, i.e. translation, rotation and scaling. Motion is created by capturing the transformations of the set of joints for a number of points in time. 

In previous work, and in common motion file formats (e.g., BVH), the orientation of each joint is expressed based on the coordinate axes of its parent, i.e., in local coordinates. Since the human body is characterized by strong joint hierarchies, it can be treated as a kinematic chain starting from the root. In fact, it has been found that the use of some sort of relative coordinates improves the stability of the network and makes the learnt frames reusable in the 3D space \cite{Vemulapalli:2016}. In this paper, we adopt a similar configuration: we treat the hip as the root joint and express each joint's position and rotation according to the root. Thus, we predict the orientation and displacement of each joint with respect to the root, which we refer to as \textit{current coordinate system}. We model the root displacement as a separate component. 
% We, thus, utilize a \textit{current coordinate system} for each joint, which makes the representation invariant to global transformations. 
We can express the current transformation of the root joint using local homogeneous coordinates of the form:
% the root using homogeneous coordinates of the form:
%
\begin{equation}M_{curr,root} = 
\begin{bmatrix}
r_{11} &r_{12}&r_{13}&0\\
r_{21} &r_{22}&r_{23}&0\\
r_{31} &r_{32}&r_{33}&0\\
0&0&0&1
\end{bmatrix}.
\end{equation}
% \na{clarification here: for the root the displacement can be anything. We define it to be the offset of the root, which usually is either 0,0,0 or 0,height,zero. In our implementations, it is 0,0,0 for the CMU because the root offset is 0,0,0 and 0,height,0 for the Mixamo.}
Then following the tree hierarchy and using the local homogeneous coordinates of each joint $j$, \begin{equation}M_{loc,j} = 
\begin{bmatrix}
r_{11}  & r_{12} & r_{13} & \text{offset}_x\\
r_{21}  & r_{22} & r_{23} & \text{offset}_y \\
r_{31}  & r_{32} & r_{33} & \text{offset}_z\\
0       & 0      & 0      & 1
\end{bmatrix},
\end{equation} 
we can compute the current homogeneous representation for each joint using: 
\begin{equation}
M_{curr,j} = M_{curr,(j-1)} \times M_{loc,j}
\end{equation}

Obtaining the local rotations w.r.t. each joint's parent in the tree architecture, as well as the offsets is straightforward when animation files are used. Intuitively, we can recover the local rotation of joint $j$ using the inverse procedure:
\begin{equation}
M_{loc,j} = M^{-1}_{curr,(j-1)}M_{curr,j}
\label{eq: curr2loc}
\end{equation} 
This representation allows us to proceed similarly to traditional techniques, in order to produce the animation in the desired format.

\begin{figure}[t]
    \centering
    \includegraphics[width=\linewidth]{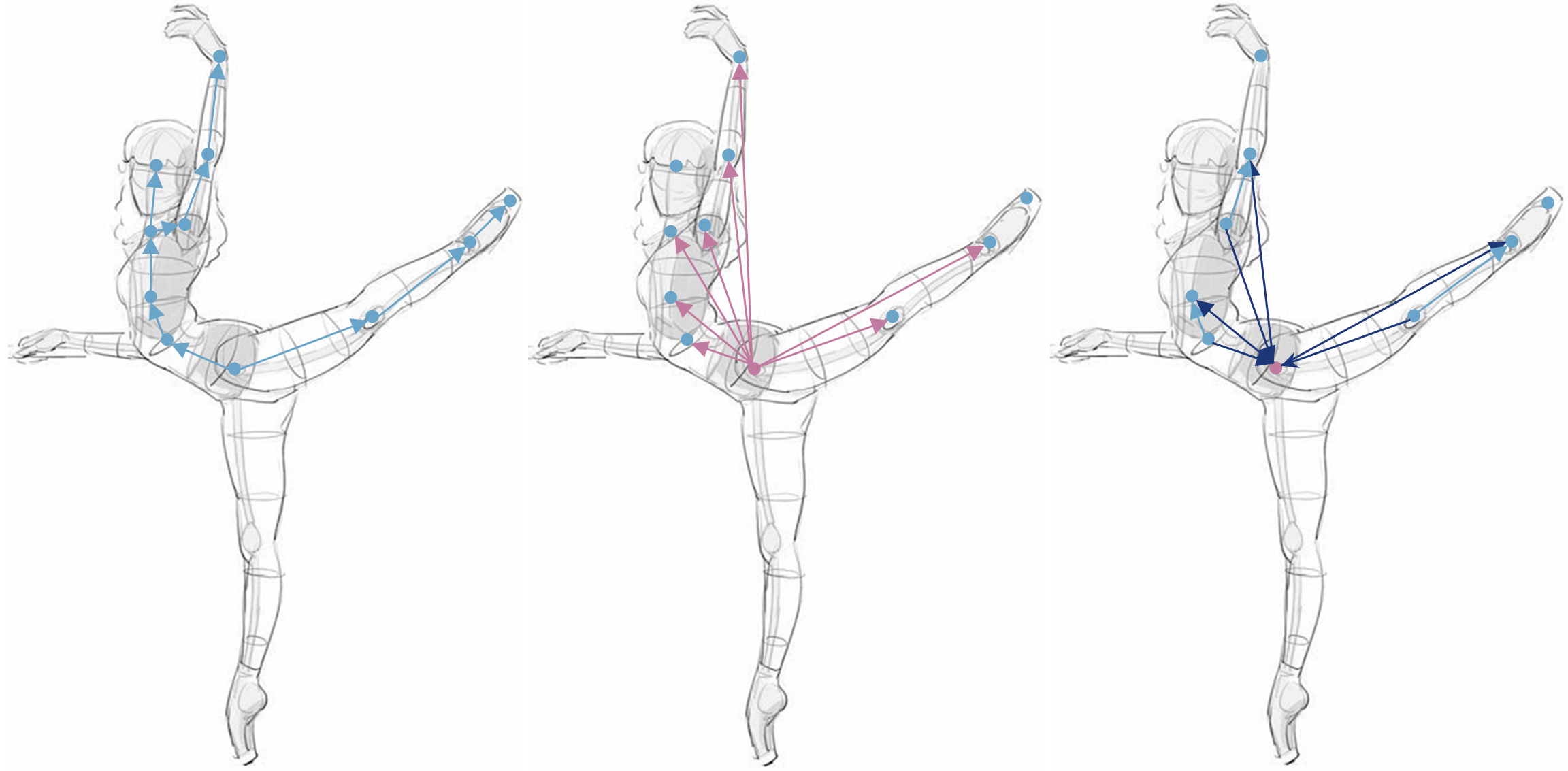}
    \caption{From left to right: local coordinate system, current coordinate system (purple), the procedure of retrieving the local orientation given the current orientations (dark blue). In our representation, we encode the rotation and translation components in current coordinates, i.e. with respect to the root joint.}
    \label{fig:lossdistance}
\end{figure}

Joint current configuration offers many advantages over the commonly used local representation. That is because during inference we estimate the orientation of each joint independently - relative to a root joint - making our predictions less vulnerable to accumulated errors from local orientations. This process is illustrated in Figure~\ref{fig:lossdistance}. Our encoding, which is untangled in the next section, handily allows us to recover local coordinates and orientations.

\section{Dual Quaternion Motion Representation}
\label{sec:Dual Quaternion Motion Representation}
An ideal representation would be one which incorporates prior skeletal information, and allows the extraction of precise information on the orientation of each joint. The use of transformation matrices would result in a $4 \times 4$ representation for each joint of the skeleton, leading to a high-dimensional representation which may contain excessive information. Instead, such information can be encoded in a dual quaternion.
% simultaneous positional and rotational information can be encoded into dual quaternions which allow for smooth interpolation between frames and are not prone to the gimbal lock effect. 

\subsection{Dual Quaternion Encoding}
\label{Section: Dual Quaternion Encoding}
Setting the dual quaternion for the root equal to 
$\ubar{\mathbf{q}} = (w_r + x_r\mathbf{i} + y_r \mathbf{j} + z_r\mathbf{k}) + \varepsilon(0 +0\mathbf{i} +0\mathbf{j} + 0\mathbf{k}) \label{eq:1}$ and the dual quaternion of each joint as $\ubar{\mathbf{q}} = (w_r + x_r\mathbf{i} + y_r \mathbf{j} + z_r\mathbf{k}) + \varepsilon(w_d +x_d\mathbf{i} +y_d\mathbf{j} + z_d\mathbf{k})$ we can construct the hierarchical representation using the dual quaternion operations, equivalent to those of transformation matrices. The rotation coefficients $w_r,x_r,y_r,z_r$ can be obtained by conversion of local Euler angles to quaternions (see Eq.~\ref{eq:rot}), while the displacement coefficients $w_d,x_d,y_d,z_d$ can be obtained using Eq.~\ref{eq:displacement}.

\begin{figure}[t]
    \centering
    \includegraphics[width=\linewidth]{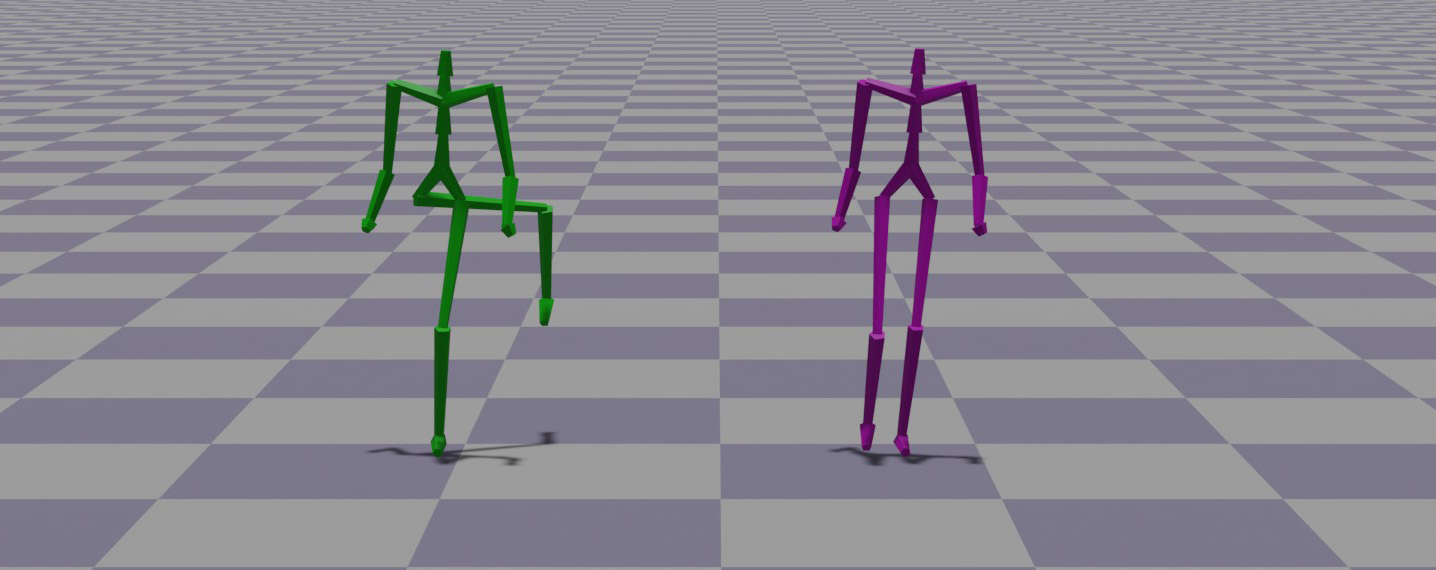}
    \caption{Abnormal motion generated with data for which we do not correct for the antipodal property during preprocessing (green) compared to the expected motion (purple).}
    \label{fig:antipodal}
\end{figure}

% The above procedure can be performed using dual quaternions instead of transformation matrices. The equivalent of matrix multiplication is dual quaternion multiplication ~\cite{Kenwright:2013}. The equivalent of the homogeneous matrix inverse is the combined conjugate i.e. $(\mathbf{q}_r,\mathbf{q}_d) = (\mathbf{q}_r^*,-\mathbf{q}_d^*)$, where $\mathbf{q}_r^*,\mathbf{q}_d^*$ are the quaternion conjugates. It is crucial to make sure that unit dual quaternions are used. 
Dual quaternions, just like ordinary quaternions, are known to exhibit the antipodal property. That is, $\mathbf{\ubar{\hat{q}}}$ and $-\mathbf{\ubar{\hat{q}}}$ represent the same rigid transformation. We tackle this phenomenon by adopting, in preprocessing, the technique employed in Kavan~\etal~\cite{Ladislav:2008} and Pavllo~\etal~\cite{Pavllo:2018}. Among $\mathbf{\ubar{\hat{q}}}$ and $-\mathbf{\ubar{\hat{q}}}$, we choose the representation with the lowest Euclidean distance from the representation of the previous frame. We apply this step for both dual quaternions as well as quaternions. As shown in Figure~\ref{fig:antipodal}, bypassing this step led to uneven interpolation, of the form of instantaneous flickering and global shaking in the generated motion. To make sure that the generated dual quaternions remain meaningful and retain their dual nature, we explicitly normalize them using the following formula:
\begin{align}
   \hat{\mathbf{\ubar{q}}} &= \frac{\ubar{\mathbf{q}}}{||\ubar{\mathbf{q}}||} =   \frac{\mathbf{q}_r}{||\mathbf{q}_r||} + \varepsilon \bigg[ \frac{\mathbf{q}_d}{||\mathbf{q}_r||} -  \frac{\mathbf{q}_r}{||\mathbf{q}_r||}  \frac{<\mathbf{q}_r , \mathbf{q}_d>}{||\mathbf{q}_r||^2} \bigg] \nonumber \\
   &= \mathbf{\hat{q}}_r +  \varepsilon \bigg[ \frac{\mathbf{q}_d}{||\mathbf{q}_r||} - \mathbf{\hat{q}}_r  \frac{<\mathbf{q}_r , \mathbf{q}_d>}{||\mathbf{q}_r||^2} \bigg]
\end{align}

\subsection{Dual Quaternion Learning}
\label{Section: Losses}
Dual quaternions not only form a unified and compact representation, but also eliminate the need for a forward kinematics layer which is commonly used to recover positions and employ a corresponding loss~\cite{Pavllo:2018, Zhou:2019, Shi:2020}. We have carefully defined the dual component to encode the 3D orientation of each joint relative to the root (current coordinates), % so that we can utilize the maximum potential of dual quaternions by 
and incorporated the following losses:
% to employ a positional loss. Since positional information is implicitly encoded into the dual quaternion representation, we can directly extract the translation component from the representation itself, using dual quaternion algebraic properties. We can then employ the following losses: 

\begin{enumerate}
    \item If we consider a dual quaternion as an 8-dimensional vector, then we can employ the Mean Squared Error (MSE) between the predicted and generated dual quaternions, given by:
    \begin{equation}
    \mathcal{L}_{\text{mse}} =\| \mathbf{dq} - \mathbf{\tilde{dq}}\|_2^2
    \end{equation}
    where $\mathbf{dq}, \mathbf{\tilde{dq}}\in \mathbb{R}^{8}$ denote the predicted and true dual quaternion for one joint. 

    \item \textbf{Rotational Loss} is the error between predicted and ground-truth rotations. The loss can be calculated in Euler space or quaternion space, similar to Kavan~\etal~\cite{Ladislav:2008}. We can measure this loss on the % current or 
    local rotations. %The current rotations are taken directly as the first 4 values of the dual quaternion representation for each joint.
    To recover the local rotations we perform the relevant operations along the hierarchy to recover the local dual quaternions, and then extract the local quaternions. % In both cases, we convert quaternions to Euler angles as discussed in Section \ref{section: Motion Representation}. Since Euler angles wrap around $2\pi$, we compute the L1 norm modulo $2\pi$ between predicted ($\mathbf{\tilde{w}}$) and ground truth ($\mathbf{w}$) Euler angles using:
%    \begin{equation}
%        \mathcal{L_{\text{euler}}}  = \|(\mathbf{w} - \mathbf{\tilde{w}} + \pi) \text{ mod } 2\pi  - \pi\|_1
%    \end{equation}.
%    This loss does not account for the continuity of prediction. Instead, 
    We can measure the error using the following equation:
    \begin{equation}
        \mathcal{L_{\text{quat}}} = 1 - \mathbf{q} \cdot \mathbf{\tilde{q}}
        \label{eq:L_quat}
    \end{equation}
    where $\cdot$ denotes the dot product between ground-truth and predicted quaternions $\mathbf{q}$ and $\mathbf{\tilde{q}}$, respectively.
    
    \item \textbf{Positional Loss:} The main benefit of the dual quaternion representation is the capability to directly extract the local position of each joint %. Using the mathematical framework defined in Section \ref{section: Motion Representation} we extract the positional information 
    ($\mathbf{p} \in \mathbb{R}^{3}$). \na{Since we have adopted  the current coordinate system, and utilizing the algebra of dual quaternions we can get the position of each joint using Eq.~\ref{eq:translation}}.% We then compute the per joint Euclidean distance between predicted positions and ground-truth positions. 
    For each joint the loss $ \mathcal{L_{\text{pos}}}$ is given by computing the Euclidean distance between predicted and ground-truth positions \na{, i.e. $
    \mathcal{L_{\text{pos}}} =\| \mathbf{p} - \mathbf{\tilde{p}}\|_2
    $
    %\begin{equation}
    %\mathcal{L_{\text{pos}}} = \sqrt{\sum_{i=1}^{3}(p_i - \hat{p}_i)^2}.
    %\mathcal{L_{\text{pos}}} = \|\mathbf{p} - \mathbf{\tilde{p}}\|.
    %\end{equation}
    where $\mathbf{\tilde{p}}$ is the predicted joint positions.}
    
    \item \textbf{Offset Constraint:} To maintain that no skeletal violations occur, we explicitly pose skeletal restrictions on the joint offsets ($\mathcal{L_{\text{offset}}}$). This can be achieved by recovering the offset of each joint as the translation encoded in a local dual quaternion. \na{To get that information from our representation we first convert the current dual quaternions to local using \ref{eq: curr2loc} and then extract the positional information (which corresponds to the offset $\mathbf{o}\in\mathbb{R}^3$ in the local coordinate system) using Equation~\ref{eq:translation}}. The loss is measured as the Euclidean distance between the predicted and ground truth offset\na{, i.e.   
    $\mathcal{L_{\text{offset}}} =\| \mathbf{o} - \mathbf{\tilde{o}}\|_2$
    where $\mathbf{\tilde{o}}$ is offset extracted from the predictions.}
    
    \item \textbf{Penalty Loss (Regularization Term):} To stabilize the learning process we include a regularisation term in the loss ($\mathcal{L_{\text{reg}}}$), responsible for preserving the unity conditions of the dual quaternion (similarly applied for quaternions). It is calculated prior to normalization of the generated frames to reflect the fact that only unit dual quaternions represent a valid transformation. This loss is based on the unitary conditions (Equation \ref{eq:unit}), i.e. 
    \begin{equation}
    \begin{split}
        \mathcal{L_{\text{reg}}} &= (w_r^2+x_r^2+y_r^2+z_r^2-1)^2 \\ & + (w_rw_d + x_rx_d + y_ry_d + z_rz_d)^2
        \end{split}
     \end{equation}

\end{enumerate}

% To stabilize the learning process we include a regularisation term in the loss, responsible for preserving the unity conditions of the dual quaternion (similarly applied for quaternions). It is calculated prior to normalization of the generated frames to reflect the fact that only unit dual quaternions represent a valid transformation. This loss is based on the unitary conditions (Equation \ref{eq:unit}. 

\section{Experimental Setup}
\label{section: Experimental Setup}
In this section, we explain the experimental setup used to evaluate the performance of our pose representation. In our experiments we utilize two popular architectures, the acRNN by Zhou~\etal~\cite{Zhou:2018}, and QuaterNet by Pavllo~\etal~\cite{Pavllo:2018}\na{; we chose these networks%Even though these networks are not the state-of-the-art currently, we chose to work with them 
since they are simple recurrent networks which process sequential data, but do not involve complex structures and inductive biases which might influence the performance. The proposed representation encourages the learning of both spatial as well as temporal correlations in recurrent architectures, i.e. while the RNN focuses on the temporal context, the representation implicitly encodes the spatial context. %Our selection of networks is also based on the fact that in both setups the root displacement is separated from the pose, making it convenient to apply our representation. 
Keeping the networks fixed, we assess the performance of different pose representations.} 
% which achieve state-of-the-art results in motion prediction and synthesis conditioned on past poses.
We design our experiments around the prediction task, since we want to assess the quality of the pose representation, without influences from other controls such as end-effector constraints, terrain information, etc. It is worth mentioning that QuaterNet, apart from motion prediction, allows for motion synthesis given a trajectory control. These architectures allow us to effectively show that the examined hierarchy-aware encoding can not only replace the FK layer, but also lead to more stable long-term synthesis via the encoding of prior skeletal information in the representation itself. It is important to note that our representation is model-agnostic, and can be incorporated in other architectures, assuming that the kinematic tree (hierarchy and offsets) are known.
% redundant: In this work, we revisit the performance of other popular representations perform several experiments and an ablation study on the losses.

Our experiments have been conducted on a PC with NVIDIA GeForce 2080 Ti GPU, and Intel Core i9-9980HK at 2.4 GHz CPU with 32GB RAM. The deep learning frameworks used are implemented using PyTorch. We used data taken from the Carnegie Mellon University motion capture database~\cite{CMU:2021} (subjects 61, 86) for the acRNN network. Motion files were originally captured at 120 frames per second, but were subsampled to 30 frames per second without much loss of temporal information. We split the dataset in training, validation, and testing sets using the 75/10/15 \% configuration. The exact details, \na{as well as the weights used for different losses,} can be found in the Appendix. For the QuaterNet we use the exact same dataset and configuration as in the original paper. %More details can be found in Table \ref{tab:acLSTM configuration} in the Appendix. 
Training the acRNN network takes about 6.5 hours (around 1500 epochs) when using positional input, 9 hours when the quaternions, quaternions-positions and ortho6D are used, and 11 hours for our representation and ortho6D-positions.
% \na{and 9 for ortho6D}.
On the other hand, training QuaterNet short-term takes 20 minutes for 3000 epochs using quaternions, and 45 minutes for dual quaternions, and QuaterNet long-term takes 2 hours for 4000 epochs using quaternions, and 3.5 hours for dual quaternions.
% \na{The conversion between quaternions to dual quaternions in the short-term experiment takes approximately 7 minutes and 4 minutes for the long-term.}
All visualisations have been created using the publicly available rendering engines of Aberman~\etal~\cite{Aberman:2020:Retargetting}. \na{The code used for the transformations between quaternions and dual quaternions, local and current dual quaternions, as well as all the code related to algebraic operations and losses of dual quaternions can be found on our project page}.  % that exhibits the aforementioned characteristics.

% \subsection{Experimental Setup}
% \label{subsection: Experimental Setup}
In our experiments, human motion is represented as a combination of $J$ joint coordinates for which we describe the orientation using various parameterizations. In addition, the global translation is modelled using the Cartesian coordinates of the root joint.

%relative to the hip, along with the root displacement from its position in the previous frame, in global space. 

\subsection{acRNN}
This network is capable of synthesizing complex human motion with the use of recurrent neural networks. The model consists of 3 LSTM layers with a memory size of 1024, followed by a linear layer. During training, the network receives as input a sequence of 100 frames, consisting of a mixture of ground truth frames from the database and frames predicted by the model, and then generates the consecutive 100 frames. During test time, it receives a seed motion of 10 frames with the goal of producing extended sequences of complex human motion. Training is performed in batches 32, optimized using Adam~\cite{Kingma:2015} with learning rate 0.0001. 

In the original implementation of acRNN, human motion is represented with joint positions relative to the root, and training is performed using the 
%which is paired with 
MSE loss that measures the similarity between the predicted and ground-truth frames. We modify the original network to receive as input: (a) local quaternions, (b) the ortho6D representation~\cite{Zhou:2019}, and (c) current dual quaternions. The quaternion and ortho6D representations are used to parameterize local rotations, while the dual quaternions operate using the current rotations and positions. Note that, similarly to our representation, quaternions can also be further constrained (in addition to the MSE loss) with the rotational ($\mathcal{L}_{quat}$) and positional losses ($\mathcal{L}_{pos}$). Recall that, the latter requires an additional FK layer, which converts local rotations to positions; we refer to this combination as quaternions-FK. Finally, to distinguish whether the performance gain in the learning is a result of the use of dual quaternions, or the combination of positions and rotations, we assess two more input formats: (a) local quaternions with current positions, appended for each joint, and (b) the ortho6D representation with the current positions. In that case, we employ only the MSE loss. In the following sections, we will refer to these two representations as quaternions-positions and ortho6D-positions.
% that is is additionally constrained with an extra FK layer that the positions  and 

% \Nefeli{mention ortho6D with FK requires constructing the rotation matrix}
% In contrary, the dual quaternion representation operates using the current rotations and positions. 

% For cases (a) and (c) we examine various losses to assess their impact. To be precise, in addition to the MSE loss, for local quaternions we experiment with the $\mathcal{L}_{quat}$ loss, as described in Section~\ref{Section: Losses}, and the forward kinematic loss on current positions, as proposed in previous work~\cite{Villegas:2018,Pavllo:2018,Aberman:2020:Retargetting}. We refer to this combination as quaternions-FK. Unless otherwise stated, quaternion refers to the simple case, trained with MSE only. 

The network receives as input the motion sequence $\mathbf{m}_t \in \mathbb{R}^{3 + DJ}$, where the first 3 entries of each sequence contain the root translation, and $D$ is the dimension of each representation ($J = 31$, $D = 3$ for positions, $D = 4$ for quaternions, $D = 6$ for the ortho6D, $D = 8$ for dual quaternions, $D = 7$ for quaternions with positions, and $D = 9$ for ortho6D with positions). Note that, during training we  normalize the features of the acRNN network by subtracting the mean and dividing by the standard deviation of motion of the training set to account for different scales in the feature space and speed up the optimisation. 

%human motion is represented as a combination of $J=31$ local joint positions relative to the hip, along with the velocity of the root in global space. We modify the original acRNN to receive as input: (a) unit quaternions; and (b) dual quaternions. \AAA{To mitigate foot-sliding errors, we incorporate foot contact labels, $\mathbf{c}_t \in \{0,1\}^2$, which are extracted during preprocessing and are calculated based on the distance of joints from the ground.} That is, network receives as input the motion sequence $\mathbf{m}_t \in \mathbb{R}^{100 \times (3 + DJ + 2)}$, where the first 3 entries of each sequence contain the root displacement from its position in the previous frame, and $D$ is the dimension of the representation ($D = 3$ for positions, $D = 4$ for quaternions, and $D = 8$ for dual quaternions). \Andreas{we need to explain what the number 100 is referring to}

\subsection{QuaterNet}
We decided to experiment with QuaterNet since it accomplishes both short- and long-term motion modeling, and uses local quaternions as the pose representation. %We utilize the recurrent architecture to explore both short-term prediction and long-term generation using the proposed motion representation. 
The short-term task concerns predictions over small time intervals of the future. Instead of predicting absolute rotations, at each frame, quaternion multiplication is used to obtain rotation deltas, while learning is based on the angle error between the predicted and ground-truth rotations. On the other hand, the long-term task is defined as the generation of motion that is conditioned on control variables, which specify the trajectory and speed. In this case, future motion is predicted using absolute rotations, and learning is achieved by integrating a positional loss that requires FK to convert the rotations to positions. 
% Short-term evaluation can be achieved by measuring the similarity between the predicted and ground truth motion.
%To further explore the performance of the proposed hierarchical representation, 

In this experiment, we modify the original QuaterNet architecture to additionally take as input dual quaternions. For a fair comparison, we train the original and adapted networks using the exact same datasets, losses, and number of epochs. In particular, for training the short-term task we use the local rotational loss on the Euclidean space. For the long-term task, we use the positional ($\mathcal{L_{\text{pos}}}$), offset ($\mathcal{L_{\text{offset}}}$), and rotational ($\mathcal{L_{\text{quat}}}$) losses. %\na{here mention that this is the original framework and we keep the losses fixed, it is not a choice that we have made. 
Instead of employing FK on the rotations, we derive the predicted positions directly from the output representation, circumventing the need for additional kinematic operations. Note that, long-term motion generation is highly uncertain and difficult to evaluate quantitatively as no ground-truth exists. For that reason, we assess the visual quality of the generated motion compared to that generated with ordinary quaternions. Finally, for both short- and long-term tasks, the global displacement of the root is controlled through the trajectory and speed parameters, inferred by the so-called pace network, which is a simple recurrent network. %\na{However, instead of being inferred, these high-level control features can be provided by the user, allowing for controlled synthesis.}

\section{Results and Evaluation}
\label{section: Experiments and Evaluation}
In the following section, we present our findings for each of the two networks. Note that,  motion shaking cannot be illustrated in snapshots, thus refer to our supplementary video for animated results. 

\subsection{acRNN}
\subsubsection{Ablation Study}

We begin our experiments by evaluating the variants of our representations in the acRNN architecture, using two motion subjects from the CMU dataset. The details for the train/validation/test configuration can be found in Table~\ref{tab:acLSTM configuration}. Through an ablation study, we are able to evaluate the impact and the effectiveness of each of the proposed losses. The first experiment consists of the motion of subject 86, which consists of locomotion and other simple motions. The second experiment is performed on more complex motion, i.e. salsa dancing (subject 60-61). 

First, we examine the loss used in the original implementation by Zhou~\etal~\cite{Zhou:2018}, namely the $\mathcal{L_{\text{mse}}}$. 
  % First, we experiment with the acRNN network. 
We found that $\mathcal{L_{\text{mse}}}$ dominates over all other losses proposed when used simultaneously. Therefore, to examine the effect of the remaining loss components, which are specific for our representation, we removed the MSE loss in this experiment. %(note that, in our experimental setup, similarly to the original paper, we only use the MSE loss). 
We neglect the hip position and only focus on the quality of generated poses. We notice that the use of the positional loss alone ($\mathcal{L_{\text{pos}}}$) leads to ambiguity in the learning. This is reflected in abnormal generated poses. However, by additionally incorporating the offset loss ($\mathcal{L_{\text{offset}}}$), we observe that the poses are corrected, with minor artifacts on the hands. Finally, by incorporating the rotational loss, $\mathcal{L_{\text{quat}}}$, we observe that the generated motion remains natural and realistic on longer duration, with no artifacts. Motion does not freeze, however, it might get repetitive from a certain point onward; we believe that is because of the nature of the architecture, and not the representation itself, and can be resolved with more training. %Note that for this experiment we adjusted the learning rate to account for the scale of each loss. 

% By incorporating the positional ($\mathcal{L_{\text{pos}}}$) and offset ($\mathcal{L_{\text{offset}}}$) losses, we observe that motion stagnates early \na{or gets repetitive, } and does not recover, indicating that these losses alone do not assist the network to learn patterns which allow long-term synthesis. 
% since the motion stagnates from an early point. 
% \Andreas{what did we have observed? shaking movement, abnormal rotations, jittery joints with absurd changes in direction? Be specific. Its good, as soon as we have the rendering, to explicitly describe a case in a figure}. 
% By additionally incorporating $\mathcal{L_{\text{quat}}}$ on the local rotations, we observe that the network is capable to learn, and generates natural and realistic motion of longer duration. Note that, in these experiments we ignore the global root displacement since it is not related to the representation. The incorporation of the MSE loss allows for higher quality and variability of the generated motion, and enables long-term generation of up to 30000 frames of motion. Motion does not freeze, however gets repetitive from a certain point onward; \na{we believe} that is because of the nature of the architecture, and not the representation itself.

% \na{We observed that when the motion is simple, the positional, offset and rotational losses provide similar results to those generated using the MSE loss alone. 

We observed that when the motion is simple, the MSE loss, $\mathcal{L_{\text{mse}}}$, is sufficient to produce similar results compared to those obtained by including all other losses together. However, on data with more complex and dynamic movements, the addition of the remaining losses ($\mathcal{L_{\text{quat}}}$, $\mathcal{L_{\text{offset}}}$, $\mathcal{L_{\text{pos}}}$), can further improve the quality of motion, eliminating minor artifacts present. Incorporating various losses, though, requires investigation to adjust the weights assigned to each loss. However, for our representation the MSE loss is sufficient for smooth motion to be produced, even for 300000 frames. 

\subsubsection{Results and Evaluation}

\na{In terms of qualitative experiments,} firstly, we evaluate the performance of the different quaternion-based representations: the quaternions with MSE only, the quaternions-FK, and the quaternions-positions. For this experiment we use the CMU dataset, subject 86. Among the three representations, we find that the quaternions-FK is more stable, with less shaking, especially on the feet; our findings are inline with previous works~\cite{Pavllo:2018,Villegas:2018,Aberman:2020:Retargetting}. We also observed that appending the positions with quaternions does not improve the generated motion, leading to increased shaking. We believe that this happens since the two components are learnt independently, and there is no constraint forcing them to match.

In a similar setup, we compare the ortho6D representation with ortho6D-positions. In this case, we were unable to clearly distinguish among the two, since abnormalities were only minor in both cases. We believe that appending the current positions in the input did not provide improvements,  while increasing the complexity of the model (see  Table~\ref{tab:acLSTM params} of the Appendix).

\begin{table*}[h]
    \centering
    \caption{\na{Quantitative evaluation: (a) NPSS ($\downarrow$), (b) Euclidean Loss ($\downarrow$) on joint positions, and (c) the acceleration.}}
    \small
    \begin{tabular}{l||c|c|c||c|c|c||c||}
     & \multicolumn{3}{c||}{NPSS ($\downarrow$)} &  \multicolumn{3}{c||}{Euclidean ($\downarrow$)} & Acceleration ($\downarrow$) \\
     \hline
     & \multicolumn{3}{c||}{Duration in $\mu s$} &  \multicolumn{3}{c||}{Duration in $\mu s$} & Duration in $\mu s$ \\
     representation    & $100$ & $400$ & $1000$ & $100$ & $400$ & $1000$ & $1 \times 10^4$\\
     \hline
     \hline
    quaternions (MSE)        & 0.49 & 0.70 & 1.19 & 2.29 & 2.28 & 2.41 & 0.18 \\
    quaternions-FK           & 0.38 & 0.52 & 1.03 & 1.96 & 1.98 & 2.19 & 0.10 \\
    quaternions-positions (MSE)   & 0.50 & 0.69 & 1.15 & 2.09 & 2.09 & 2.22 & 0.25 \\
    ortho6D (MSE)      & 0.42 & 0.55 & 0.86 & 1.93 & 1.87 & 2.05 & 0.13 \\
    ortho6D-positions (MSE) & 0.40 & 0.48 & 0.81 & 1.57 & 1.57 & 1.79 & 0.11 \\
    dual-quaternions (MSE)   & 0.44 & 0.53 & 0.87 &1.62  & 1.63 & 1.86 & 0.10 \\
    dual-quaternions (all)   & \textbf{0.30} &\textbf{ 0.36 }&\textbf{ 0.70} & \textbf{1.36} &\textbf{1.46} & \textbf{1.78} & \textbf{0.08}
    \end{tabular}   
    \label{tab:metrics1}
\end{table*}

\begin{figure}[t]
  \centering
    \includegraphics[width=\linewidth]{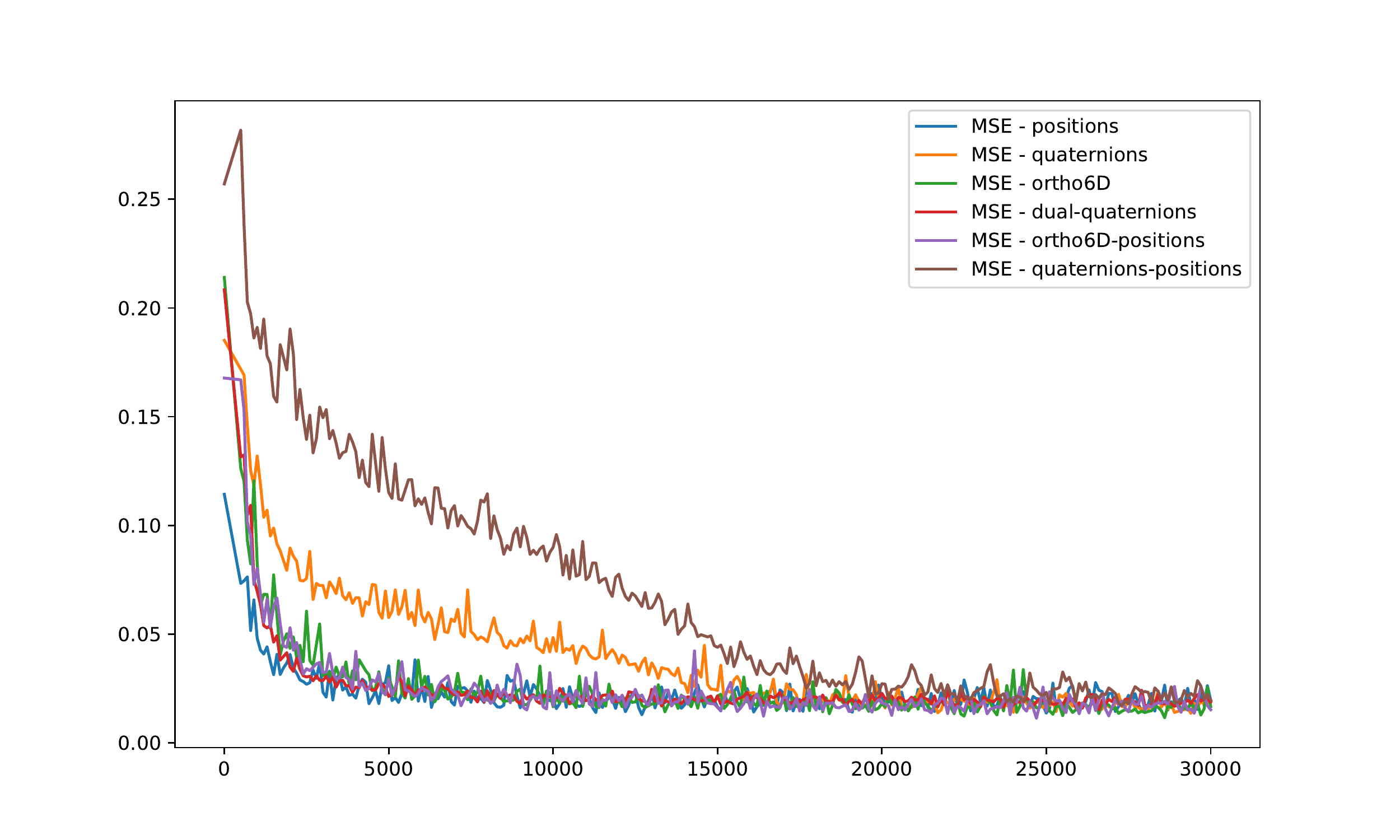}
  \caption{Training and validation losses for each representation. Loss curves converge faster for dual quaternion and positional compared to quaternions.}
    \label{fig:acLSTMloss}
\end{figure}

Our next experiment focuses on the learning and convergence. Figure~\ref{fig:acLSTMloss} shows the training loss for the positional, ortho6D, quaternion, with their variants, and our representation. It can be seen that quaternion, and its variants, have the slowest convergence. We believe that this is due to the well-studied quaternion interpolation problem~\cite{Zhang:2018,Zhou:2019}. Directly regressing on positions is indeed effective and converges fast, at the cost of ambiguity when recovering the rotations. On the other hand, 
ours and the ortho6D representations have similar convergence: ortho6D avoids the quaternion interpolation problems by using the forward and upward vectors, while our hierarchy-aware encoding is richer in information, thus facilitating learning. 

We qualitatively assess the generated motion at various learning stages. %We compare our representation, with the best variants of ortho6D and quaternions, as identified earlier: the local quaternions, ortho6D, and ours. 
In this setup, we only use the MSE loss, for the sake of fair comparisons, among all three representations: the local quaternions, ortho6D, and ours. We generate results after training for 30K, as well as, 150K iterations. With 30K training iterations, we observe that the results generated using the quaternion and ortho6D representations are shaky and exhibit artifacts at certain times, while with our representation the generated motion is stable. With 150K iterations, the artifacts observed using the other two representations are reduced, yet, minor shaking is still evident in the generated motion. On top of that, we compare our representation, incorporating all losses, with the best variant of quaternions (quaternions-FK), and the ortho6D representations. Again, we train each network for 150K iterations, and observe that those minor artifacts, even though reduced, are still present when using quaternions-FK and ortho6D, but are not evident when using ours. Please refer to the video for an animated visual illustration.

Finally, we perform an additional experiment on the CMU salsa dataset, to assess the ability of our representation in synthesizing long-term sequences of complex motion (up to 30K frames). The generated motion does not freeze, however, gets repetitive from a certain point onward; that is because of the nature of the architecture, and not the representation itself. We compared our results with the corresponding quaternions and ortho6D, and observed that the quality of motion generated using our representation is more stable.

\na{We employ several metrics to measure the quantitative performance of our representation in comparison to the baselines. In particular, we measure the frame-wise Euclidean distance and the Normalized Power Spectrum Similarity (NPSS)~\cite{Gopalakrishnan:2019}; NPSS measures the similarity of power spectrum between ground truth and the corresponding generated sequence. %Based on Gopalakrishnan~\etal~\cite{Gopalakrishnan:2019}, the squared magnitude spectrum of the Discrete Fourier Transform coefficients $X_{i,j}[f],Y_{i,j}[f]$, are first normalized as $X^{norm}_{i,j}[f] = \frac{X_{i,j}[f]}{\sum_f X_{i,j}[f]}; Y^{norm}_{i,j}[f]= \frac{Y_{i,j}[f]}{\sum_f Y_{i,j}[f]}$. Then NPSS is calculated as $NPSS = \frac{\sum_i \sum_j p_{i,j}*emd_{i,j}}{\sum_i \sum_j p_{i,j}}$ where $emd_{i,j} = \left\|X^{norm}_{i,j}[f] - Y^{norm}_{i,j}[f]\right\|_1, p_{i,j} = \sum_f X^{norm}_{i,j}[f]$ and $\left\|.\right\|_1$ is the L1 norm. 
Both metrics are calculated on joint positions, which are computed using Forward Kinematics. During this calculation, the global root displacement is ignored (zero) since we want to focus our evaluation on the pose. We evaluate all models at 150000 iterations. Following the original evaluation~\cite{Zhou:2018} we sample 400 seed motions (10 frames) with a sliding window of 7 frames from the test set and generate motions of different intervals $\{100,400,1000\} \mu s$. As demonstrated in Table~\ref{tab:metrics1}, our pose representation achieves the lowest scores.}

%\begin{table}[h]
%    \centering
%     \begin{tabular}{c|c|c|c}
%     & \multicolumn{3}{c}{Duration $\mu s$}\\
%     \hline
%     representation    & 100 & 400 & 1000 \\
%     \hline
%     \hline
%    quaternions & 0.49 & 0.70 & 1.19\\
%    quaternions-FK & 0.38 & 0.52 & 1.03\\
%    quaternions-positions & 0.50 &  0.69 & 1.15\\
%    ortho6D & 0.42& 0.55 & 0.86\\
%    ortho6D-positions & 0.40 & 0.48 & 0.81\\
%    dual-quaternions &  \textbf{0.30} &\textbf{ 0.36 }&\textbf{ 0.70}
%    \end{tabular}   
%    \caption{NPSS ($\downarrow$) on joint positions}
%    \label{tab:metrics1}
%\end{table}

%\begin{table}[h]
%    \centering
%     \begin{tabular}{c|c|c|c}
%        & \multicolumn{3}{c}{Duration $\mu s$}\\
%        \hline
%         representation    & 100 & 400 & 1000 \\
%        \hline
%        \hline
%    quaternions & 2.29 & 2.28 & 2.41\\
%    quaternions-FK & 1.96 & 1.98 & 2.19\\
%    quaternions-positions & 2.09 & 2.09 & 2.22\\
%    ortho6D &  1.93& 1.87 & 2.05\\
%    ortho6D-positions & 1.57& 1.57 & 1.79\\
%    dual-quaternions & \textbf{1.36} &\textbf{1.46} & \textbf{1.78}
%    \end{tabular}   
%    \caption{Euclidean Loss ($\downarrow$) on joint positions}
%    \label{tab:metrics2}
%\end{table}}

\na{Furthermore, we calculate the acceleration as an indication for the amount of jitter. We expect the mean acceleration of the predictions to be as close as possible to the real data. All models are evaluated at 150000 iterations and the root is ignored. We sample 400 seed motions (10 frames) with a sliding window of 7 frames from the test set and generate 300 frames, corresponding to 10 seconds of motion. As can be seen in Table~\ref{tab:metrics1}, our representation constrained with all losses has the lowest acceleration error.}

\subsection{QuaterNet}
\subsubsection{Ablation Study}

We perform a separate ablation study for the QuaterNet architecture. In this setup, we evaluate the contribution of $\mathcal{L_{\text{pos}}}$, $\mathcal{L_{\text{offset}}}$, and $\mathcal{L_{\text{quat}}}$ on local and current rotations, by removing one loss at a time and re-training the model. Note that, in this experiment we do not examine the contribution of the MSE loss, since it is not part of the original implementation. 
%two rotationals 
We observe that the use of either the $\mathcal{L_{\text{quat}}}$, local or current, allows for the generation of smooth motion when applied on the current dual quaternion.
%\na{remove: It is important to note that the use of a rotational loss based on Euler angles results in unexpected flickering of motion at sparse intervals, since the mapping between the representation space and the original motion is not continuous~\cite{Zhou:2019}.  
%Experiments with the rotational loss indicate that they cannot encode the critical positional constraints induced when a positional loss is placed.} 

Removing $\mathcal{L_{\text{quat}}}$, does not significantly affect the quality of the overall motion. However, certain artifacts occur in the end-effectors, particularly on hands. We believe that they occur because the positional loss alone does not explicitly constrain the end-effector's orientation.
% we constrain up to the last joint on the rig and orientation of the the end-effectors is not properly constrained.

% Potentially, they occur because we infer the positional data and penalize for offset violations, rather than performing FK on the ground truth offsets in order to penalize for rotations. We have found that \na{higher fidelity and elimination of these artifacts can be achieved by combining the positional, offset, and rotational losses to fully constrain the generated dual quaternions.} \na{(verify) Compared to the motion generated using quaternions with FK, the motion generated using dual quaternions presents less foot-sliding artifacts.} Please refer to the supplementary video for the visual results.
% overall motion remains smooth and realistic, but certain artifacts occur in the end-effectors. we found that the end-effectors, hands in particular, present some artifacts. 
Furthermore, we re-train our network without integrating $\mathcal{L_{\text{offset}}}$. In this scenario, we found that the positional loss, $\mathcal{L_{\text{pos}}}$, alone is not sufficient to guarantee that the generated motion remains realistic. In 
the original implementation, the positional loss alone was sufficient since FK were used to retrieve the positional data given the ground-truth offsets. In contrast, with our representation we need to explicitly ensure that no skeletal violations occur. This can be achieved by additionally integrating the offset loss $\mathcal{L_{\text{offset}}}$. In fact, we conclude that $\mathcal{L_{\text{offset}}}$ is a significant component in the hierarchy-aware encoding. 
% Please refer to the supplementary video for animated results. 
Note that, we observe that the use of bone length constraint (i.e. incorporating a loss that  measures the differences in the ground truth and predicted bone lengths), instead of the proposed offset loss, fails to preserve the nature of skeletal structure.

% Using only $\mathcal{L_{\text{pos}}}$ and $\mathcal{L_{\text{offset}}}$, though, the end effector rotations present artifacts. By additionally adding $\mathcal{L_{\text{quat}}}$, the network leads to similar visual results as those obtained using the original implementation, but with the extra FK layer. %constrained by the positional loss (after employing the FK layer). 
%  \Andreas{Rotational loss alone was tested - explain that this does not encode the critical positional constraints that our method extracts from the hierarchy topology.}% However, by adding 
% We additionally re-train the model with only

% we evaluated the performance of our representation by adding bone length constraints \na{, i.e. incorporating a loss that measures the differences in the ground truth and predicted bone lengths}, but the results indicate that this constraint alone is

\subsubsection{Results and Evaluation}
QuaterNet is used to quantitatively compare the results generated using the original quaternion and dual quaternion representation for short-term tasks. The numerical evaluation reveals the superiority of our hierarchy-aware encoding for longer time predictions. As seen in Table~\ref{tab:short-term}, our representation obtains lower rotational error, up to 65\% for predictions of duration more than 1000ms, whereas we achieve a marginally higher error for predictions on shorter time intervals. The better performance on longer duration indicates that our implicit hierarchical encoding learns elements of the skeletal hierarchy, which prevent the motion from diverging far from a reference point. It is important to clarify that, while joint rotation accuracy is an evidence that the model can generate natural motions, our major evaluation is based on the naturalness of the generated motion as can be seen later in the long-term setup.
% \na{repetitive? The  that is present in the representation  t, particularly for longer time intervals.} 
% \na{not convincing enough... The lower performance for shorter time intervals can be explained as a result of the higher dimensionality of the proposed representation.} We further examine our claim by incorporating our representation in the long-term task framework. 
%
\begin{table}[t]
    \centering
    \small
    \caption{Average errors over all actions on test set in Euler space for short-term task on subject 5.}
    % \na{Quantitative evaluation beyond 4000ms is not applicable due to the stochastic nature of human motion. }}
    \label{tab:short-term}
    \begin{tabular}{c|c|c}
      Time (ms) & Quaternions   & Dual Quaternions \\
      \hline
      \hline
        80      &   0.3871   &          0.4091      \\
        160     &   0.6766   &          0.7250      \\
        320     &   1.0117   &          1.1086      \\
        400     &   1.1452   &          1.2612      \\
        600     &   1.4327   &          1.5686      \\
        800     &   1.7983   &          1.8111      \\
        1000    &   2.3004   &  \textbf{2.0584}     \\
        2000    &   6.0033   &  \textbf{2.8334}     \\
        3000    &  10.7987   &  \textbf{3.8045}     \\
        4000    &  13.1485   &  \textbf{4.7124}     \\
    \end{tabular}
\end{table}

% \Andreas{We need to discuss our observations about long term, short term, and explain why our method has better results. What are the results for QuaterNet when (a) quaternions are used, rotational loss? (b) quaternions with FK is used, positional loss (c) dual quaternions are used, our losses. Comment visually! Also, we have to show that our representation improves the quality of generated motion for longer periods of time by partially eliminating unwanted rotations, and also, we have to demonstrate that instead of performing FK to recover the positional information of each joint, we can directly extract such information from the representation itself.}

\na{Similarly to Pavllo~\etal~\cite{Pavllo:2018}}, we measure the performance of our pose representation in terms of the quality of the generated motion using the long-term setup.
% We consider the generated motion using dual quaternions and positional loss during training as the baseline. %In the original paper, the skeleton is parameterized with 26 joints for the long term task. It is worth mentioning that we have performed experiments with a reduced skeleton encoded in our hierarchical representation, however in that case special care should be taken in encoding joints which belong to the basis of the hierarchy. 
We verify the findings of Pavllo~\etal~\cite{Pavllo:2018}, by training the QuaterNet using quaternions, with and without employing a positional loss. In this latter setup, the generated motion is uncontrolled and suffers from perceptual abnormalities at certain points. That is because averaging the rotation errors by assigning equal weights in all joints, fails to embed that errors in crucial joints of the hierarchy, significantly affect the resulting pose.
% does not encode the importance of crucial joints in the final pose. 

% only the rotational loss does not properly constrain motion, allowing errors in crucial joints. 
When incorporating the positional loss, our observations are inline with the findings of Pavllo~\etal~\cite{Pavllo:2018}, namely, that the motion produced remains smooth and perceptually correct. This is due to the fact that the positional loss better constrains the motion, even though it allows for occasional mistakes on rotations which are not visible to the eye. Our representation produces similar results to the quaternion with the FK layer, since it encodes positional information within. \na{Please refer to the supplementary video for animated experimental results.}
% e observed that the quaternion representation without the positional loss results in abnormal motion of the end-effectors, especially on turns.

% \subsection{Ablation Study}
% \label{sec:ablation}
% To evaluate the impact and the effectiveness of each of the proposed losses, we conduct an ablation study. %by training each network with different losses. 
% \subsubsection{acRNN}

% \subsubsection{QuaterNet}

\subsection{Applications}
% Usually, when training models for motion processing tasks, all data points are retargeted to a common universal skeleton. 
One of the main advantages of our method is that it encodes features dependent on the skeletal proportions, allowing training on skeletal-variant datasets, without requiring that they are a priory retargeted onto a universal model. We leverage upon the fact that the network can learn correlations between motion patterns and skeletal attributes through the hierarchy-aware encoding. We examine whether these can be transferred onto unseen motions and skeletons. To do so, we use the acRNN architecture and the MIXAMO dataset. We compare our representation with 2 other input representations: (a) quaternions with offsets appended, and (b) ortho6D with offsets appended. We train the models for approximately 170 epochs. We compare with those two representations which explicitly take into consideration skeletal aspects of the different characters. For all representations, we use the MSE loss for the fairness of comparison. Our aim is to highlight that the skeletal features are implicitly incorporated into the representation and assist learning. 

% the case of quaternions-offsets and ortho6D-offsets, we use the MSE loss. For our representation, we incorporate the the rotational $\mathcal{L_{\text{quat}}}$, positional $\mathcal{L_{\text{pos}}}$ and offset $\mathcal{L_{\text{offset}}}$ losses. }

% For the case of quaternions, we use the MSE loss $\mathcal{L_{\text{mse}}}$ and the local rotational loss $\mathcal{L_{\text{quat}}}$. For our representation, we additionally incorporate the positional $\mathcal{L_{\text{pos}}}$ and offset $\mathcal{L_{\text{offset}}}$ losses. 
We limit our experiments to training on 4 distinct skeletons (Abe, Michelle, Timmy, and James), which have similar skeletal topology, but different proportions. We split the motion for each character into training/validation and testing, and consider the following 2 motion prediction scenarios: (a) seen character, but unseen motion; and (b) unseen character, but seen motion. We use a different character (Mousey) as the unseen skeleton. Finally, to further test the capabilities of our representation, we test it for the scenario of unseen character and unseen motion. Note that unseen motion refers to motion which was not directly used during training, but a contextually similar one could have. 

% In this experiment we aim to investigate this claim and explore whether the dual quaternion representation is well-suited for learning from a training set of heterogenous skeletons.

With the original evaluation setup, i.e. feeding 10 motion frames as seed to initialize the generation, and various skeletons present in the training database, we have observed that the network's capabilities in synthesizing long-term motion decrease. Thus, for qualitative comparisons, we slightly adapt the evaluation protocol. We feed a sequence of 100 ground-truth frames ($f_t : f_{t+100}$), and generate a total of 250 frames: the first 100, ($f_{t+1}:f_{t+101}$) are based on ground-truth input frames, and the rest are generated based on the network's outputs. This allows us to have a reference motion for visual comparisons. Note that due to sub-sampling, some motions end up having less than 100 frames. In such cases, motion is generated based on ground-truth frames, for the maximum number of available frames. Once these are exhausted, predictions are generated based on the network's predictions.

% \Andreas{what is the experiment and its scope? We need to describe it... Also, we need to give brief details of our implementation, the baseline implementation (e.g., quaternions), the losses used . See Kfir's paper, section 5.1}. %, which combines a variety of motion for characters with different morphologies. 
\begin{figure}[t]
    \centering
    \includegraphics[width=\linewidth]{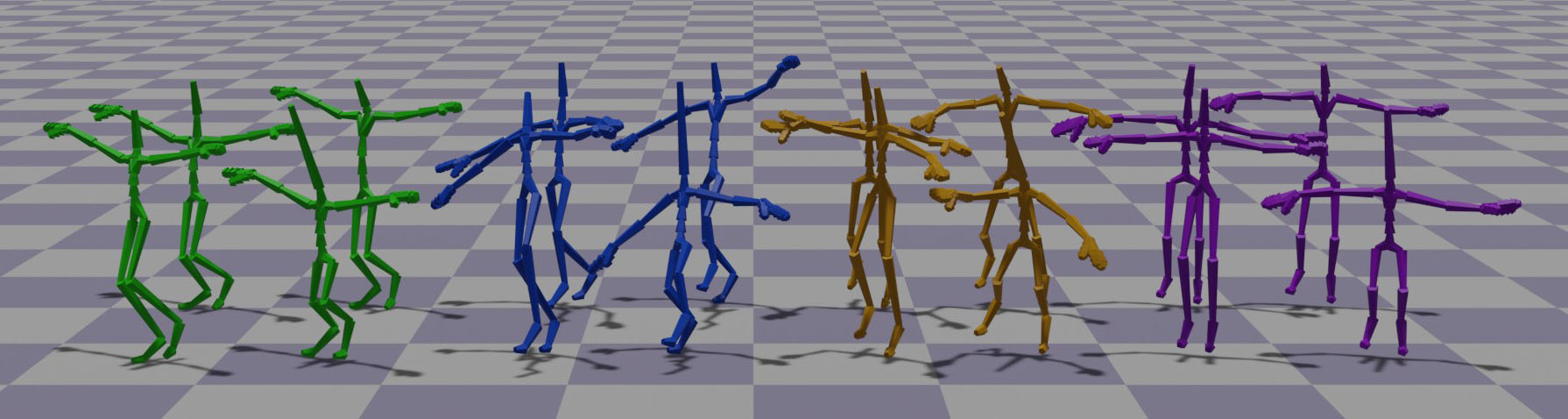}
    \caption {Static screenshot of motion generated with unseen seed motion on seen characters. The green group depicts the seed motion, the blue represents motion generated using quaternions-offsets, the orange represents motion generated using ortho6D-offsets, and the purple motion generated with our representation.}
    \label{fig:applications_all}
\end{figure}

\begin{figure}[t]
    \centering
    \includegraphics[width=\linewidth]{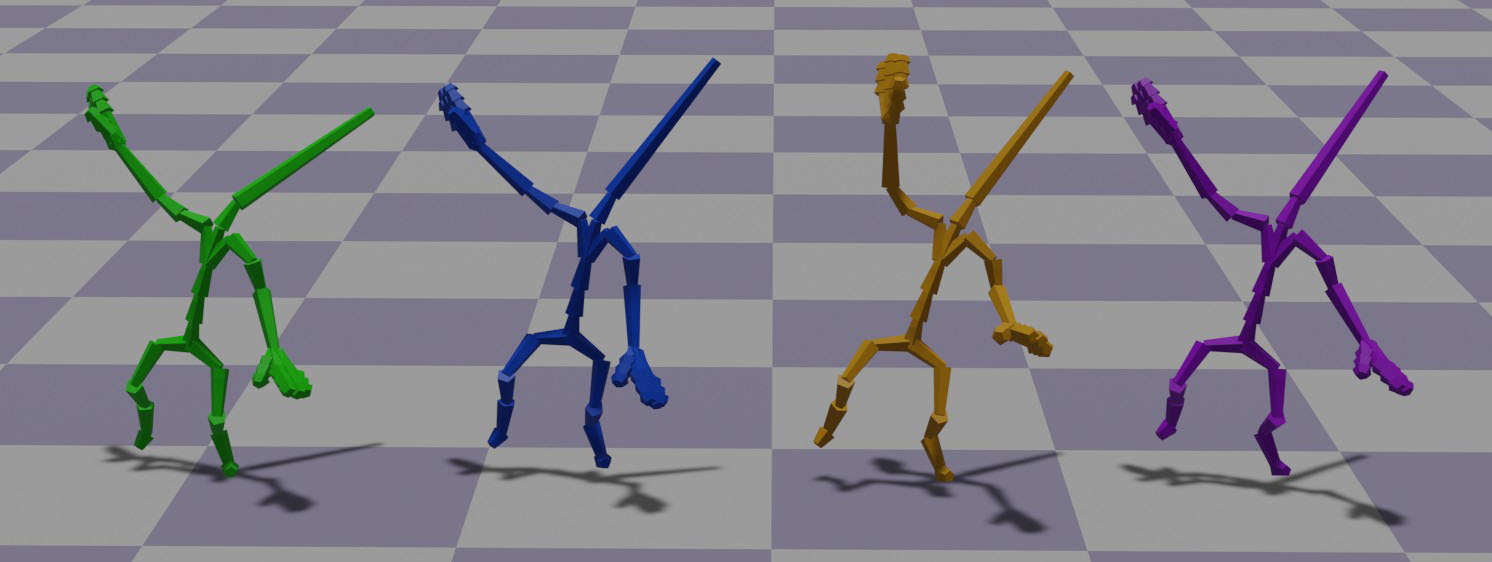}
    \caption {The seen motion, shown in green, is transferred to an unseen character. The pose of the blue character (quaternion-offsets) and orange character (ortho6D-offsets) are abnormal, whereas the purple (ours) is closer to the reference motion.}
    \label{fig:applications_unseen_char_seen_mot}
\end{figure}

% We only predict the 25 most important joints of each skeleton which are common for the characters used.
In all the examined scenarios, our representation shows the ability to satisfy perceptually sensitive constraints by encoding prior information about the hierarchy and skeletal topology. This results to motion prediction that more closely resembles the reference motion of each character. In contrast, when motion is represented using quaternions-offsets and ortho6D-offsets, the predicted motion is shaky, with jittering on several joints. Note that, in our visualizations we neglect the root translation, that is not part of our representation, and needs to be treated differently for various skeletons. Figure~\ref{fig:applications_all} shows the generated motion on characters from the training set, initialized with unseen seed motion. Figure~\ref{fig:applications_unseen_char_seen_mot} illustrates motion generated on an unseen character, with seed seen motion. As can be seen in the accompanying video, the motion generated with our representation is the most robust. % The case of seen motion on unseen character is of particular interest, since it shows that motion of another character seen during training can be transferred to an unseen character with similar morphology. 
Our representation is capable of producing smoother motion, even in the extreme scenario of unseen motion on unseen character. Our findings give a promising direction for the field of deep skeletal animation, revealing that the encoding of skeletal information into the representation, lays the grounds for training on skeletons with varying proportions. Please refer to the accompanying video for the animated results.
% instead of retargeting to a common skeleton and loosing skeleton-specific motion attributes.

%  \AAA{Fig.~\ref{} shows... It can be observed that our method ... compared to the use of unit quaternions, where...} 

% \Andreas{Use some if the following sentences:  
% \begin{itemize}
%     \item our method is closely resembling the original source motion. 
%     \item Our pose representations satisfies perceptually sensitive constraints on the joint position predictions, due to the fact that the network learn the skeletal hierarchy topology,  which are typically achieved by applying IK optimizations.
%     \item when unit quaternion is used as pose representation, the generated motion is shaky, with absurd jittering on several joint rotations 
%     \item Figure~\ref{} shows snapshots from the comparison. It can be seen that our results (right column) are more stable than when unit quaternions are used and aligned better with the ground truth (a green skeleton, overlaid on top of each output), in terms of global positioning and local errors of joints.
% \end{itemize} 
% }

\section{Conclusion}
\label{section: Conclusion}
%Our work is paving the way towards the establishment of a concrete pipeline for character animation using deep learning. The proposed hierarchy-aware parameterization 
%The proposed parameterization can be employed in any existing network. 
We have presented a skeletal pose representation, well-suited for deep skeletal animation. Our parameterization, which is the first to employ dual quaternions as the input to neural networks for motion modelling, %mathematical framework that represent pose as input to neural networks, 
learns to infer both rotational and positional information directly from the training data, resulting in a more constrained synthesis with improved realism. Results, using two different network architectures, demonstrate that our representation, which encodes the correlations between joints and limbs along the rig, enables the prediction of coherent and stable longer-term motion sequences. It also supports training on various skeletons, contrary to most works which assume a unified skeleton. Thus, the representation preserves motion attributes and nuances that are usually lost when retargeting motion onto a common skeleton for training purposes.
% transferring motion from one character to another. %during retargeting. 

%\paragraph{Limitations} 
% \na{include limitations based on reviewer's comments}
One limitation of our method relates to the way that the root translation is handled, when training with different skeletons. In this work, this information is being ignored and is corrected manually for the visualization. % In this work we focus on the pose with respect to the root joint. 
Future work could focus on handling the root translation when training with different characters. Furthermore, the encoding of both positional and rotational information in a unified representation %can be adopted for any character animation task 
comes at a small cost, computational complexity. \na{This is attributed to two factors. Firstly, during training, the network has to learn an 8-D representation for each joint compared to 3, 4, or 6 dimensions of other representations. Secondly, even though the conversion from local to current coordinates and construction of dual quaternions is performed during pre-processing, some calculations performed during training such as conversion from local to current coordinate frame or vice versa, the extraction of the positions and normalization result in marginally increased training times.} 
 %\na{include number of parameters} % for Euler angles, quaternions and the ortho6D \cite{Zhou:2019} representations which only encode rotations. \Andreas{do we have results for the ortho6D}. 
%Alternatively to dual quaternions, one could argue that transformation matrices can be used, resulting in a 16-dimensional representation per joint, twice as larger as the proposed representation. 
However, the increased complexity is not a limiting factor since the synthesis/prediction is still achieved in real-time. \na{In the future, our implementation could be further optimized to reduce the computational complexity.} Finally, we have shown the applicability of the representation in motion prediction using recurrent architectures. Future work can exploit the performance of representation in other types of architectures (feed-forward, convolutional), other applications (motion retargeting or reconstruction), controlled environments, or interactions.

% It has been shown that our representation offers great advantages over conventional methods, when training neural networks, and can find several applications in motion prediction or synthesis, motion retargeting, motion reconstruction, or any other applications that are skeleton tuned. 

% In future work, we will further examine the dual quaternion properties on various other network architectures, controlled environments, and evaluate their high transferablity factor to different applications. 
% Finally, we note that for representations we can additionally incorporate foot-contact labels and apply IK mechanism in a post-processing phase, as in previous works~\cite{Aberman:2019}. 
% even though foot artifacts are a network problem, not a problem caused by the representation, \na{explain that we didn't want to additionally include contact labels to change the performance, but these can be added} foot contact labels can be added to improve motion realism.

%\paragraph{Future work} 
%can will

% ------------------------------------
% Acknowledgements
% ------------------------------------
\section*{Acknowledgements}

The authors would like to thank Anastasios Yiannakidis for his help in rendering the results and helping with the implementation of the acRNN model, as well as Victoria Fernández Abrevaya for fruitful discussions. This work has received funding from the European Union’s Horizon 2020 research and innovation programme under the Marie Skłodowska-Curie grant agreement No 860768 (CLIPE project). This project has also received funding from the University of Cyprus, and the European Union's Horizon 2020 Research and Innovation Programme under Grant Agreement No 739578 and the Government of the Republic of Cyprus through the Deputy Ministry of Research, Innovation and Digital Policy.

% ------------------------------------
% References
% ------------------------------------
%\bibliographystyle{eg-alpha-doi}
%\bibliography{References}

\printbibliography

% ------------------------------------
% Appendix
% ------------------------------------
%\newpage
\section*{Appendix}
\label{sec:Appendix}
\renewcommand{\thesubsection}{\Alph{subsection}}

%\Andreas{You have to explain what are these tables, and cite/refer them into the text.}
\na{\subsection{Training data for acRNN}}
\begin{table}[h]
    \centering
    \small
    \caption{Train/Validation/Test split configuration for experiments with the acRNN architecture~\cite{Zhou:2018}}
    \label{tab:acLSTM configuration}
    \begin{tabular}{c|c|c|c}
        CMU Data Subject & Train & Validation & Test \\
        \hline
        \hline
        \textbf{Subject 86} & 1,2,4,5,7,8,10,12,13-15  & 6  & 3,9  \\
        \hline
        \textbf{Subject 61} & 1-12  & 13  &14,15 \\ 
    \end{tabular}
\end{table}

\na{\subsection{Loss Weights}
For the acRNN experiments, $\mathcal{L_{\text{pos}}}$ and $\mathcal{L_{\text{quat}}}$ is weighted by $\frac{1}{3}$, FK by $\frac{1}{4}$ and $\mathcal{L_{\text{quat}}}$ by 1. When only the MSE is used, the features are not weighted.  The weight of the regularizer which is used to penalize the generation of un-normalized quaternions and dual quaternions, $\mathcal{L_{\text{reg}}}$, is set to 0.01.

For the QuaterNet experiments, $\mathcal{L_{\text{reg}}}$ is weighted by 0.01, and all other losses (  $\mathcal{L_{\text{pos}}}$, $\mathcal{L_{\text{off}}}$, rotational loss) by 1.

\na{\subsection{Model Complexity}}
\begin{table}[h]
    \centering
    \small
    \caption{Number of total parameters of the acRNN model for each representation}
    \label{tab:acLSTM params}
    \begin{tabular}{r|c}
        \textbf{Representation} & \textbf{Parameters} \\
        \hline
        \hline
        Positional & 21579890  \\
        % \hline
        Euler angles & 21487712  \\
    %   \hline
        Quaternions (CMU) & 21646463  \\
        % \hline
        Quaternions (Mixamo) & 21523559  \\
        %  \hline
        Ortho6D (CMU) & 21963965  \\
        %  \hline
        Ortho6D (Mixamo) & 21779606  \\
        %  \hline
        \hline
        Dual quaternions (CMU) &  22281467 \\
        %   \hline
         Quaternions+positional (CMU) &  22122716 \\
        %   \hline
        Ortho6D+positional (CMU) & 22440218  \\
        %   \hline
        Dual quaternions (Mixamo) & 22035659  \\
        %   \hline
        Quaternions+positional/offset (Mixamo) &  21907634 \\
        %   \hline
        Ortho6D+positional/offset (Mixamo) & 22163684   \\
           \hline
        
    \end{tabular}
\end{table}
}

\end{document}